\documentclass{article}
\usepackage{ijcai24}
\usepackage{times}
\usepackage{soul}
\usepackage{url}
\usepackage[hidelinks]{hyperref}
\usepackage[utf8]{inputenc}
\usepackage[small]{caption}
\usepackage{graphicx}
\usepackage{amsmath}
\usepackage{amsthm}
\usepackage{booktabs}
\usepackage{algorithm}
\usepackage{algorithmicx}
\usepackage{algpseudocode}
\usepackage[switch]{lineno}
\usepackage{xcolor}

\usepackage{amsthm}
\usepackage{newfloat}
\usepackage{listings}
\usepackage{amsmath}

\usepackage{amssymb}
\usepackage{lipsum}
\usepackage{subcaption}
\usepackage{tabularx}
\usepackage{multirow}
\usepackage{adjustbox} % table size
\usepackage{booktabs} % top rule & midrule
\usepackage{makecell}
\usepackage{enumitem}

\usepackage{verbatim}
\usepackage{appendix}

\newcommand{\sysname}{FADE}
\newtheorem{definition}{Definition} 
\newtheorem{problem}{Problem}
\title{FADE: Towards Fairness-aware Generation for Domain Generalization via Classifier-Guided Score-based Diffusion Models}

\author{
Yujie Lin$^{1,2}$\thanks{Equal contribution.}
\and
Dong Li$^{3*}$
\and
Minglai Shao$^{1}$\thanks{Corresponding author.}
\and
Guihong Wan$^4$
\and
Chen Zhao$^3$\\
\affiliations
$^1$School of New Media and Communication, Tianjin University\\
$^2$School of Informatics, Xiamen University \\
$^3$Department of Computer Science, Baylor University\\
$^4$Departments of Biostatistics and Epidemiology, Harvard University\\
\emails
linyujie001213@gmail.com,
shaoml@tju.edu.cn,
\{dong\_li1, chen\_zhao\}@baylor.edu,
gwan@mgh.harvard.edu
}
\begin{document}
\maketitle
\begin{abstract}
Fairness-aware domain generalization (FairDG) has emerged as a critical challenge for deploying trustworthy AI systems, particularly in scenarios involving distribution shifts. Traditional methods for addressing fairness have failed in domain generalization due to their lack of consideration for distribution shifts. Although disentanglement has been used to tackle FairDG, it is limited by its strong assumptions. To overcome these limitations, we propose Fairness-aware Classifier-Guided Score-based Diffusion Models (\sysname{}) as a novel approach to effectively address the FairDG issue. Specifically, we first pre-train a score-based diffusion model (SDM) and two classifiers to equip the model with strong generalization capabilities across different domains. Then, we guide the SDM using these pre-trained classifiers to effectively eliminate sensitive information from the generated data. Finally, the generated fair data is used to train downstream classifiers, ensuring robust performance under new data distributions. Extensive experiments on three real-world datasets demonstrate that \sysname{} not only enhances fairness but also improves accuracy in the presence of distribution shifts. Additionally, \sysname{} outperforms existing methods in achieving the best accuracy-fairness trade-offs.
\end{abstract}

\section{Introduction}
\label{sec:intro}

In recent years, fairness in machine learning has received widespread attention due to its significance in real-world systems such as loan approvals~\cite{purificato2023use} and employee hiring~\cite{alder2006achieving}. From a group perspective, algorithmic fairness focuses on the statistical parity among different groups defined by specific sensitive attributes of people~\cite{hardt2016equality}. 
Many efforts have been made to address algorithmic fairness~\cite{mitchell2021algorithmic}. A type of method achieves fairness by eliminating bias from the dataset through fair data generation. As a pioneer in this approach, FGAN~\cite{xu2018fairgan} removes sensitive information by ensuring that the discriminator cannot distinguish the sensitive group membership of the generated samples. Due to the powerful generative capabilities, diffusion models have also been employed to ensure the fairness of generated data. FLDGM~\cite{ramachandranpillai2023fair} integrate an existing debiasing method with diffusion models to generate unbiased data. Although they have achieved remarkable results, their poor generalization performance has caused them to fail when facing distribution shifts~\cite{shao2024supervised}.

Fairness-aware domain generalization (FairDG), as an emerging topic, has gained increasing attention because it is essential for advancing the deployment of trustworthy AI in the real world~\cite{shao2024supervised}. 
Instead of the independent and identically distributed (\textit{i.i.d.}) assumption, this field focuses on fairness under distribution shifts. The most widely studied distribution shift is covariate shift~\cite{shimodaira2000improving}, which variations to changes caused by differences in the marginal distributions of instances. One class of methods addresses the FairDG issue by learning fair and domain-invariant representations through feature disentanglement~\cite{oh2022learning}. However, these methods often have overly strong assumptions that it can perfectly disentangle sensitive information and domain-specific information from the feature representation. The assumptions limit their applicability and reduce their robustness. Another class of methods enhances the diversity of the training set through data augmentation~\cite{pham2023fairness}, enabling the model to generalize better to unseen target domains and thereby improving overall performance.

In this paper, inspired by data augmentation strategies, we propose a three-stage framework called \sysname{}: Fairness-aware Classifier-Guided Score-based Diffusion Models to generate unbiased and domain-invariant data. Specifically, we pre-train the parameters of the score-based diffusion model (SDM), label classifier, and sensitive classifier using a source dataset across multiple domains in the first stage. This stage equips the three components with generalization capabilities in the target domain. In the second stage, we derive a generator by guiding the SDM with the pre-trained classifiers, which effectively removes bias or sensitive information from the generated data. Simultaneously, during the generation phase, we can obtain fair data without requiring any additional training. Finally, we use the generated data, which has reduced sensitive information, to train a fairness-aware domain-invariant downstream classifier. Our contributions can be summarized as follows:
\begin{itemize}
    \item 
    We formulate a novel problem: generating unbiased data to train downstream classifiers that are tested on distribution-shifted datasets, while ensuring both accuracy and fairness.
    \item We have designed a novel fair data generation method called \sysname{}, which ensures the generalization ability of the generative model during the pre-training phase and removes biased information during the generation phase. \sysname{} not only allows for the specification of generated sample categories but also possesses generalization capabilities under new data distribution shifts.
    \item Experimental results on three real-world datasets demonstrate that \sysname{} achieves the best performance in both fairness and accuracy compared to other baselines when facing the challenge of distribution shifts.
\end{itemize}

\section{Related Work}
\label{sec: Related Work}
 \begin{figure*}[!t]
    \centering   
    \includegraphics[width=
    0.8\linewidth]{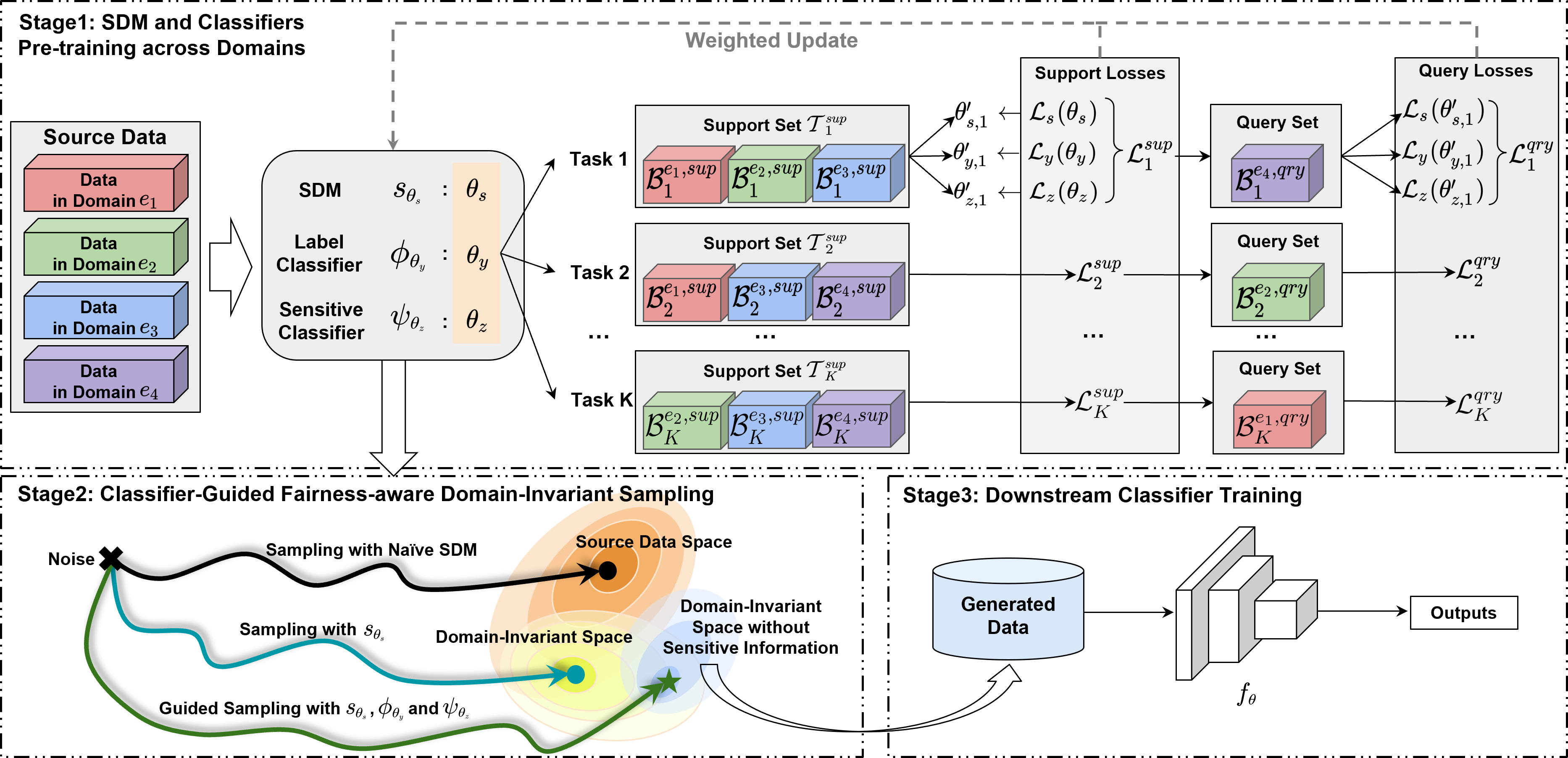}
    % \vspace{-2mm}
    \caption{ An overview of \sysname{}. 
    Supposing there are 4 domains in the training set $\mathcal{D}^{src}$. For the $k$-th iteration, we randomly select a batch as $\mathcal{T}_k^{sup}$, and sample from the remaining three domains to compose $\mathcal{T}_k^{qry}$. We perform gradient descent using the loss $\mathcal{L}^{sup}_k(\Phi)$ obtained on $\mathcal{T}_k^{sup}$ to obtain a temporary set of parameters $\Phi'$, then use the model with $\Phi'$ to obtain loss $\mathcal{L}^{qry}_k(\Phi'_k)$ values on $\mathcal{T}_k^{qry}$. We use the weighted sum of two losses to update $\Phi=\{\boldsymbol{\mathbf{\theta}}_s,\boldsymbol{\mathbf{\theta}}_y,\boldsymbol{\mathbf{\theta}}_z\}$. When using the score network $s_{\boldsymbol{\theta}_s}$ to generate data from noise, we can guide the generation process with $\phi_{\boldsymbol{\theta}_y}$ and $\psi_{\boldsymbol{\theta}_z}$ to obtain unbiased data. Finally, these data are used to train downstream tasks to achieve the final results.} 
    \label{fig:illustrasion}
    \vspace{-2mm}
\end{figure*}

\textbf{Fair Data Generation.}
Fairness in machine learning aims to ensure equitable performance across different demographic groups, and it can be achieved through three primary approaches: pre-processing, in-processing, and post-processing methods. Pre-processing methods modify the training data to mitigate biases before training the model, using techniques such as data resampling, data transformation, and fair data generation. Fair data generation 
is similar to pre-processing methods. However, unlike pre-processing methods, fair data generation does not use the original data to train downstream classifiers. Instead, it can generate additional data for training predictive models, which is especially beneficial when the original training data is very limited. FGAN~\cite{xu2018fairgan} is the first method to tackle fair data generation. Also based on GANs~\cite{goodfellow2020generative}, DECAF~\cite{van2021decaf} can achieve various fairness criteria by leveraging causal graphs. FLDGM~\cite{ramachandranpillai2023fair} attempts to integrate an existing debiasing method~\cite{liu2022fair} with GANs or diffusion models~\cite{ho2020denoising} to achieve fair data generation. 
%Unlike these existing methods, \sysname{} not only synthesizes an arbitrary number of samples but also allows for the specification of each sample's category. Additionally, models trained on data generated by \sysname{} possess the capability to handle shifts in the distribution of test data. 

\textbf{Fairness under Distribution Shifts.}
Achieving fairness is not devoid of challenges, especially in the presence of distribution shifts. These shifts can pose significant hurdles as models trained on source distributions may not generalize well to target data distributions, potentially exacerbating biases and undermining the intended fairness objectives~\cite{lin2024supervised,komanduri2024causal,lin2023towards,lin2023adaptation,zhao2019rank}. 
There are two primary approaches to addressing fairness issues across domains: feature disentanglement and data augmentation. Feature disentanglement aims to learn latent representations of data features, enhancing their clarity and mutual independence within the model~\cite{zhao2024algorithmic,lian2025metric,jiang2024feed,li2024learning,zhao2023towards}. However, these methods frequently rely on overly strong assumptions, which restrict their applicability and diminish their robustness. Additionally, assessing the quality of disentanglement presents its own challenges. Data augmentation seeks to enhance the diversity of training datasets and improve model generalization performance by systematically applying controlled transformations to the training data~\cite{pham2023fairness,zhao2024algorithmic}.

\section{Background}
\label{sec:background}
 \textbf{Notations.} Let $\mathcal{X}\subseteq\mathbb{R}^p$ denotes a feature space and $\mathcal{Z}=\{0,1\}$ is a sensitive space. $\mathcal{Y}=\{0,1\}$ is defined as an output or a label space. 
A domain $e\in\mathcal{E}$ is defined as a joint distribution $\mathbb{P}^e_{XZY}:=\mathbb{P}(X^e,Z^e,Y^e)$ on $\mathcal{H}=\mathcal{X}\times\mathcal{Z}\times\mathcal{Y}$. 
A dataset \textit{i.i.d.} sampled from a domain $e$ is represented as $\mathcal{D}^e=\{(\mathbf{x}^e_j,z^e_j,y^e_j)\}_{j=1}^{|\mathcal{D}^e|}$, where $\mathbf{x},z,y$ are the realizations of random variables $X,Z,Y$ in the corresponding spaces. 
A function $f$ parameterized by $\boldsymbol{\theta}$ is denoted as $f_{\boldsymbol{\theta}}$.
% \textcolor{red}{A classifier $f$ in the space $\mathcal{F}$ is denoted as $f:\mathcal{X}$
% $\rightarrow\mathcal{Y}$.}
We defer a list of the notations used in this paper in  Appendix \ref{sec:appendix-notaions}.

%We denote $\mathcal{E}_{src}$ and $\mathcal{E}_{tgt}$ as sets of domain labels for source and target domains, respectively.
\textbf{Fairness-aware Domain Generalization (FairDG).}
%Algorithmic fairness primarily encompasses group fairness, individual fairness, and counterfactual fairness. This paper focuses solely on the most common form, group fairness.
In the supervised learning setting of FairDG~\cite{shao2024supervised}, the goal is to lean a classifier $f_{\boldsymbol{\theta}}:\mathcal{X}\rightarrow\mathcal{Y}$ across multiple source domains $\mathcal{E}^s=\{e_i\}_{i=1}^{S}$, where $S=|\mathcal{E}^s|$, such that the learned $f_{\boldsymbol{\theta}}$ can be applied on a distinct shifted target domain $e_t\notin \mathcal{E}^s$, which is unknown and inaccessible during training, achieving both good generalization accuracy and fair decision-making.

To ensure fair predictions, the classifier $f_{\boldsymbol{\theta}}$ is required to guarantee that the outcomes $\hat{Y}=f_{\boldsymbol{\theta}}(X)$ are not biased or discriminatory against any specific protected groups, each characterized by sensitive attributes $Z$, such as race and gender. 
Specifically, a fairness-aware metric $\omega_{\boldsymbol{\theta}}:\mathcal{Z}\times\mathcal{Y}\rightarrow\mathbb{R}$, such as the difference of demographic parity~\cite{dwork2012fairness} or the difference of equalized odds~\cite{hardt2016equality}, is used to control the fair dependence between the sensitive attribute and model outcomes during training. A value of $\omega_{\boldsymbol{\theta}}(Z,\hat{Y})$ closer to $0$ indicates greater fairness.

\begin{problem}[FairDG]
Given a source dataset $\mathcal{D}^{src}=\left \{ \mathcal{D}^{e_i} \right \}_{i=1}^{S}$, where each $\mathcal{D}^{e_i}=\{(\mathbf{x}^{e_i}_j,z^{e_i}_j,y^{e_i}_j)\}_{j=1}^{|\mathcal{D}^{e_i}|}$ is \textit{i.i.d.} sampled from a unique source domain $e_i \in \mathcal{E}^s$,
% , let $\mathcal{D}^{src}=\{\mathcal{D}^s\}_{s=1}^{\left |\mathcal{E}^s \right |}$ be a set of source data and $s\in\mathcal{E}^s \subset \mathcal{E}$. 
% We have access to $\mathcal{D}^{src}$ in training process whlie the target domain $\mathcal{D}^{tgt}=\{\mathcal{D}^s\}_{t=1}^{\left |\mathcal{E}^t \right |}$ is unseen, where $t \in \mathcal{E}^t = \mathcal{E}\setminus  \mathcal{E}^s$. 
a loss function $\ell :\mathcal{Y} \times \mathcal{Y} \rightarrow \mathbb{R}$, and a fair metric $\omega_{\boldsymbol{\theta}}:\mathcal{Z}\times\mathcal{Y}\rightarrow\mathbb{R}$, the goal is to learn a fair classifier $f_{\boldsymbol{\theta}}:\mathcal{X}\rightarrow\mathcal{Y}$ that minimized the risk over $\mathcal{D}^{src}$ satisfying a fairness constraint:
% for any $D_s$ that minimizes the worst-case risk over all domains (i.e. $\left \{ \mathbb{P}_{XZY}^e\right \}_e^{\left | \mathcal{E} \right |}$):  
\begin{equation*}
    \small
    \begin{aligned}
        \min_{\boldsymbol{\theta}} \max_{e_i \in \mathcal{E}^s} \quad &\mathbb{E}_{(\mathbf{x}^{e_i},z^{e_i},y^{e_i})\in\mathcal{D}^{src},\forall e_i\in\mathcal{E}^s} [ \ell(f_{\boldsymbol{\theta}}(\mathbf{x}^{e_i}), y^{e_i}) ] \\ 
        \text {s.t.}\quad &\omega_{\boldsymbol{\theta}}\left(z^{e_i}, f_{\boldsymbol{\theta}}(\mathbf{x}^{e_i}) \right)\leq \epsilon
    \end{aligned}
\end{equation*}
where $\epsilon>0$ is an empirical threshold. The learned $f_{\boldsymbol{\theta}}$ is required to be generalizable to a target domain $e_t\notin\mathcal{E}^s$, which is unknown and inaccessible during training, such that it performs well in terms of predicted accuracy and fairness on $\mathcal{D}^{tgt}=\{(\mathbf{x}^{e_t}_j,z^{e_t}_j,y^{e_t}_j)\}_{j=1}^{|\mathcal{D}^{tgt}|}$ sampled from this domain.
\label{prob1}
\end{problem}

A key challenge in addressing Problem \ref{prob1} is determining how closely the data distributions in unknown target domains align with those in the observed source domains. 
Given the various types of distribution shifts discussed in~\cite{shao2024supervised}, this paper narrows the scope by focusing specifically on the shift between source and target domains that is solely due to covariate shift~\cite{shimodaira2000improving}, in which domain variation is attributed to disparities in the marginal distributions over input features $\mathbb{P}_X^{e_i}\neq\mathbb{P}_X^{e_t},\forall e_i\in\mathcal{E}^s$.

Inspired by existing domain generalization efforts~\cite{zhou2022domain}, data augmentation strategies, such as diffusion models~\cite{ho2020denoising}, applied to source domains enhance the diversity of the training data, thereby improving the ability of $f_{\boldsymbol{\theta}}$ to generalize to unseen target domains and ensuring domain invariance.

\textbf{Score-based Diffusion Models (SDMs).}
Denoising diffusion probabilistic models (DDPMs)~\cite{ho2020denoising} have demonstrated remarkable performance in generating high-fidelity data. 
The forward process in DDPMs gradually adds Gaussian noise to the data over a series of timesteps, eventually transforming the data into pure noise. 
% This process can be described as a Markov chain where each step adds a small amount of noise to the data. 
Let $\mathbf{x}_{0}$ be the original data, and $\mathbf{x}_{i>0},i=1,\cdots N$ be the noisy data after $i$ steps. The forward process is defined as:
{
\small
\begin{equation}
\mathbf{x}_i=\sqrt{1-\beta_i} \mathbf{x}_{i-1}+\sqrt{\beta_i} \mathbf{u}_{i-1}, \quad i=1, \cdots, N
    \label{eq:forward}
\end{equation}
}where noise $\mathbf u_{i-1}\sim \mathcal{N(\mathbf{0},\mathbf{I})}$, $\beta_i$ is a hyperparameter to control the intensity of noise addition and $N$ is the total number of noise addition times.

Score-based diffusion models~\cite{song2020score} are an alternative implementation of DDPMs. In the limit of $N \to \infty$, we define an auxiliary set of noise scales $\{\beta_i = N \beta_i\}_{i=1}^N$. 
Let $\beta\left(\frac{i}{N}\right) = \beta_i$, $\{\beta_i\}_{i=1}^N$ becomes a function $\beta(t)$ indexed by $t$. Given $t \in [0,T]$, $\mathbf{x}_t$ represents $\mathbf{x}$ at moment $t$. The forward diffusion in Eq.(\ref{eq:forward}) can be defined by a  stochastic differential equation (SDE) ~\cite{ho2020denoising}:
 {
\small
 \begin{equation}
\mathrm{d} \mathbf{x}_t=-\frac{1}{2} \beta(t) \mathbf{x}_t \mathrm{d} t+\sqrt{\beta(t)} \mathrm{d} \mathbf{w},
\end{equation}
}
where the function $\beta(t)$ is determined by the discrete hyper parameter $\beta_t$ and $\mathbf{w}$ is the standard Wiener process~\cite{zhang2018degradation}. 
Defining $h(\mathbf{x}_t)=-\frac{1}{2} \beta(t) $ and $g(t)=\sqrt{\beta(t)}
$. When generating samples, the corresponding reverse diffusion process can be described by the following system of SDEs:
{
\small
\begin{equation}
\mathrm{d} \mathbf{x}_t=\left[h(\mathbf{x}_t)-g^2(t) \nabla_{\mathbf{x}_t} \log p_t(\mathbf{x}_t)\right] \mathrm{d}\overline{t}+g(t) \mathrm{d} \overline{\mathbf{w}},
\label{eq:reverse diffusion}
\end{equation}
}where $\mathbf{w}$ is the reverse-time standard Wiener processes, and $\mathrm{d}\overline{t}$ is an infinitesimal negative time step. The score networks $s_{\boldsymbol{\theta_s}}$ is trained to approximate the partial score functions $\nabla_{\mathbf{x}_t} \log p(\mathbf{x}_t)$ and used to generate the sample features by evolving backward in time. Specially, the score loss function~\cite{song2020score} $\mathcal{L}_s:\Theta_s\times\mathcal{H}\rightarrow\mathbb{R}$ can be formulate:
{
\small
\begin{align}
    \mathcal{L}_{s}(\boldsymbol{\theta}_s,\mathcal{D}) =& \mathbb{E}_t \{ \lambda(t) \mathbb{E}_{\mathbf{x}_t(0)} \mathbb{E}_{\mathbf{x}_t(t) | \mathbf{x}_t(0)} [ \| s_{\boldsymbol{\theta}_s}(\mathbf{x}_t(t), t) \notag \\
    &- \nabla_{\mathbf{x}_t(t)} \log p_{0t}(\mathbf{x}_t(t) | \mathbf{x}_t(0)) \|_2^2 ] \},
    \label{eq:score loss}
\end{align}
}where $\lambda:[0, T] \rightarrow \mathbb{R}_{>0}$ is a positive weighting function, $t$ is uniformly sampled over $[0, T]$,
$\mathbf{x}_0 \sim p(\mathbf{x})$ and $\mathbf{x}_t \sim p(\mathbf{x}_t | \mathbf{x}_0)$. For DDPM, we can typically choose
 {
\small
 \begin{align}
      \lambda \propto 1 / \mathbb{E}\left[\left\|\nabla_{\mathbf{x}_t} \log p(\mathbf{x}_t | \mathbf{x}_0\right\|_2^2\right].
 \end{align}
 }

%\textbf{Reverse Process} aims to start from pure noise and gradually denoising it to recover the original data sample. This is achieved by training a Markov chain that learns to predict the noise that was added at each time step of the diffusion process, and then subtracts this predicted noise to move towards the data distribution. Our method FADE will generate domain-independent and sensitive-information-free data from both forward and reverse processes.

\textbf{Problem Setting.}
To address the problem of FairDG, given data $\mathcal{D}^{src}=\{\mathcal{D}^{e_i}\}_{i=1}^{S}$, where each $\mathcal{D}^{e_i}$ is sampled from a unique source domain $e_i\in\mathcal{E}^s$, the key of this paper is to seek a generator $G:\mathcal{X}\times\mathcal{Z}\times\mathcal{Y}\rightarrow\mathcal{X}\times\mathcal{Z}\times\mathcal{Y}$, initialized by a pre-trained SDM. 
A dataset generated by $G$, denoted $\mathcal{D}^{gen}=\{\mathbf{x}'_j,z'_j,y'_j\}_{j=1}^{|\mathcal{D}^{gen}|}$, can then be used to learn a downstream classifier $f_{\boldsymbol{\theta}}$ by minimizing the risk and satisfying algorithmic fairness, $\omega_{\boldsymbol{\theta}}(Z',f_{\boldsymbol{\theta}}(X'))\leq\epsilon$, over $\mathcal{D}^{gen}$. 
{
\small
 \begin{align}
 \label{eq:ourproblem}
      \min_{\boldsymbol{\theta}} \mathbb{E}_{(\mathbf{x}^{e_i},z^{e_i},y^{e_i})\in\mathcal{D}^{src},\forall e_i\in\mathcal{E}^s} [ \ell(f_{\boldsymbol{\theta}}(\mathbf{x}'), y') ],
 \end{align}
}where the sample $(\mathbf{x}',z',y')\in\mathcal{D}^{gen}$ is generated by $G$ using $(\mathbf{x}^{e_i},z^{e_i},y^{e_i})$.
During inference, the learned $f_{\boldsymbol{\theta}}$ is expected to perform well in terms of accuracy and model fairness on $\mathcal{D}^{tgt}$, which is sampled from a target domain $e_t\notin\mathcal{E}^s$ and is unknown and inaccessible during training.

% \textcolor{red}{To address Eq.(\ref{eq:ourproblem}), a novel framework \sysname{} is introduced in the following section, which consists of two learning processes: (1) the generator pre-training and (2) learning the generator.}

%Our goal is to ensure that the model trained on $\mathcal{D}_{gen}$ is fair for any downstream tasks (specifically classification tasks in this paper) in the target domain $\mathcal{D}^{tgt}$. In other words, the ultimate goal is to train a classifier $f_{\theta}$ parameterized by $\boldsymbol \theta$ using $\mathcal{D}_{gen}$, such that $f_{\theta}$ meets DP (Demographic Parity) and EOp (Equal Opportunity) criteria when classifying $\mathcal{D}^{tgt}$.
% Given source domain data $\mathcal{D}^{src}$, we aim to train a generator $G(\cdot)$ to generate a dataset ${D}_{gen}$. The sample features X in ${D}_{gen}$ should be irrelevant to domain information, only containing semantic information used for prediction, and X should not contain sensitive information. We directly use $\mathcal{D}_{gen}$
%   for downstream tasks, where only the basic loss needs to be trained without adding any additional algorithms. $\mathcal{D}_{gen}$ can be expressed as:
%   {
% \small
% \begin{equation}
%     \mathcal{D}_{gen}=G(z), z\sim \mathcal{N}(0,\mathbf{I}).
% \end{equation}
% }Therefore, Prob.\ref{prob1} can be transformed into finding a generative function $G(\cdot)$ that can generate de-biased and domain-invariant data, given Gaussian noise. In that case, we can refer to this problem as generalized and fair data generation.

\section{Methodology}
\label{sec:methodology}

We propose a novel framework, \sysname{}, to address Eq.(\ref{eq:ourproblem}) as shown in Figure~\ref{fig:illustrasion}. 
This framework has three stages. 
In the first stage, the parameters of the SDM, label classifier, and sensitive classifier are pre-trained using the source dataset across various domains. 
In the second stage, the generator $G$ is derived by guiding the SDM using classifiers with the learned parameters.
Finally, in the last stage, the data generated from $G$ with mitigated sensitive information are used to learn a fairness-aware domain invariant downstream classifier $f_{\boldsymbol{\theta}}$.

% \textcolor{blue}{[Need to update.]}FADE is to generate a dataset with sensitive and domain-specific information. Any downstream classifier trained on this data can achieve FairDG. We first introduce how to generate samples based on the scored-based diffusion model (SDM). We propose using the classifier guidance method to generate a label-balanced dataset and remove sensitive information that may appear during the sampling process. Finally, during training, a meta-learning-based approach is adopted to simultaneously adjust the SDM and the two classifiers used for guidance.

\subsection{SDM and Classifiers Pre-training across Domains}
As described in Figure \ref{fig:illustrasion}, given the source data $\mathcal{D}^{src}$ consisting of $S$ training domains, we consider $K$ tasks. Each task $\mathcal{T}_k$ is associated with a support and a query set.
The support set $\mathcal{T}_k^{sup}=\cup_{e_i\in\mathcal{T}_k^{sup}}\mathcal{B}_k^{e_i,sup}$ contains $S-1$ data batches, where each batch $\mathcal{B}_k^{e_i,sup}$ is sampled from the corresponding $\mathcal{D}^{e_i}\in\mathcal{D}^{src}$ and the query set $\mathcal{T}_k^{qry}=\cup_{e_i\in\mathcal{T}_k^{qry}}\mathcal{B}_k^{e_i,qry}$ is sampled from the remaining distinct data subset in $\mathcal{D}^{src}$.

Inspired by \cite{finn2017model}, this stage aims to learn a good parameter $\Phi$ over the $K$ tasks that can be generalized on all source domains. 
The pre-training model contains three components: a SDM $s_{\boldsymbol{\mathbf{\theta}}_s}$ introduced in the previous section, a label classifier $\phi_{\boldsymbol{\mathbf{\theta}}_y}:\mathcal{X}\rightarrow\mathcal{Y}$, and a sensitive classifier $\psi_{\boldsymbol{\mathbf{\theta}}_z}:\mathcal{X}\rightarrow\mathcal{Z}$.
To optimize $\Phi=\{\boldsymbol{\mathbf{\theta}}_s,\boldsymbol{\mathbf{\theta}}_y,\boldsymbol{\mathbf{\theta}}_z\}$ across domains, the objective function is given in Eq.(\ref{eq:pre-training}).
{
\small
\begin{align}
\label{eq:pre-training}
    &\min_{\Phi} \frac{1}{K}\sum_{k=1}^K \Big( \delta \mathcal{L}_k^{sup}(\Phi)+(1-\delta)\mathcal{L}_k^{qry}(\Phi'_k) \Big)
\end{align}
}
where
{
\small
\begin{align}
\label{eq:where}
    \Phi&=\{\boldsymbol{\mathbf{\theta}}_s,\boldsymbol{\mathbf{\theta}}_y,\boldsymbol{\mathbf{\theta}}_z\}\quad\text{and}\quad\Phi'_k=\{\boldsymbol{\mathbf{\theta}}'_{s,k},\boldsymbol{\mathbf{\theta}}'_{y,k},\boldsymbol{\mathbf{\theta}}'_{z,k}\}; \nonumber\\
    \boldsymbol{\mathbf{\theta}}'_{s,k} &= \boldsymbol{\mathbf{\theta}}_{s}-\xi_s\nabla_{\boldsymbol{\mathbf{\theta}}_s}s_{\boldsymbol{\mathbf{\theta}}_s}(\mathcal{T}_k^{sup}); \nonumber\\
    \boldsymbol{\mathbf{\theta}}'_{y,k} &= \boldsymbol{\mathbf{\theta}}_{y}-\xi_y\nabla_{\boldsymbol{\mathbf{\theta}}_y}\phi_{\boldsymbol{\mathbf{\theta}}_y}(\mathcal{T}_k^{sup}); \\
    \boldsymbol{\mathbf{\theta}}'_{z,k} &= \boldsymbol{\mathbf{\theta}}_{z}-\xi_z\nabla_{\boldsymbol{\mathbf{\theta}}_z}\psi_{\boldsymbol{\mathbf{\theta}}_z}(\mathcal{T}_k^{sup});  \nonumber\\
    \mathcal{L}^{sup}_k(\Phi) &= \mathcal{L}_s(\boldsymbol{\mathbf{\theta}}_s,\mathcal{T}_k^{sup})+\mathcal{L}_y(\boldsymbol{\mathbf{\theta}}_y,\mathcal{T}_k^{sup})+\mathcal{L}_z(\boldsymbol{\mathbf{\theta}}_z,\mathcal{T}_k^{sup}); \nonumber\\
    \mathcal{L}^{qry}_k(\Phi'_k) &= \mathcal{L}_s(\boldsymbol{\mathbf{\theta}}'_{s,k},\mathcal{T}_k^{qry})+\mathcal{L}_y(\boldsymbol{\mathbf{\theta}}'_{y,k},\mathcal{T}_k^{qry})+\mathcal{L}_z(\boldsymbol{\mathbf{\theta}}'_{z,k},\mathcal{T}_k^{qry}). \nonumber
\end{align}
}

In Eq.(\ref{eq:pre-training}), $\xi_s,\xi_y,\xi_z>0$ are learning rates and $\delta\in[0,1]$ is an empirical weight balancing support and query losses. $\mathcal{L}_s$ is the score loss function defined in Eq.(\ref{eq:score loss}). $\mathcal{L}_y:\Theta_y\times\mathcal{H}\rightarrow\mathbb{R}$ and $\mathcal{L}_z:\Theta_z\times\mathcal{H}\rightarrow\mathbb{R}$ are cross-entropy loss functions for predicting class labels and sensitive attributes, respectively.
Notice that $\boldsymbol{\mathbf{\theta}}'_{s,k},\boldsymbol{\mathbf{\theta}}'_{y,k},\boldsymbol{\mathbf{\theta}}'_{z,k}$ are updated using one or few gradient steps for rapid optimization.

After sufficient iterations, $\Phi$ becomes optimized and invariant to domains. 
This occurs because each task can be viewed as a domain generalization problem within a sub-condition where the support sets act as ``source data" and the query sets as ``target data".

Furthermore, in the second stage of \sysname{}, the SDM is guided to the generator $G$ by the label and sensitive classifiers using samples in $\mathcal{D}^{src}$, while keeping their parameters $\boldsymbol{\mathbf{\theta}}_s,\boldsymbol{\mathbf{\theta}}_y,\boldsymbol{\mathbf{\theta}}_z$ unchanged.

\subsection{Guiding the SDM to $G$ with Classifiers}
\textbf{Debiasing with Fair Control.} To mitigate sensitive information from samples during the reverse diffusion process, we propose a novel sampling strategy. Assuming a strength signal $a_{fair}\in\left [ 0,1 \right ]$, we should sample from the conditional distribution $p(\mathbf{x}_t| a_{fair}=\lambda_z)$, hyperparameter $\lambda_z$ represents the strengh of fair control. Consequently, we need to solve the conditional reverse-time SDE:
  {
\small
\begin{equation}
\mathrm{d} \mathbf{x}_t=\left[h(\mathbf{x}_t)-g^2(t) \nabla_{\mathbf{x}_t} \log p(\mathbf{x}_t| a_{fair})\right] \mathrm{d}\overline{t}+g(t) \mathrm{d} \overline{\mathbf{w}}.
\label{eq:reverse diffusion4}
\end{equation}
}Since $p(\mathbf x_t| a_{fair}=\lambda_z)\propto p(\mathbf x_t)p(a_{fair}=\lambda_z | \mathbf x_t)$,  we can derive the gradient relationship (proved in Appendix \ref{sec:appendix-proof}) as follows:
{
\small
\begin{equation}
    \begin{aligned}
    \nabla_{\mathbf{x}_t} \log p(\mathbf{x}_t| a_{fair}=\lambda_z)=&\nabla_{\mathbf{x}_t} \log p(\mathbf{x}_t)\\&+\nabla_{\mathbf{x}_t} \log p(a_{fair}=\lambda_z|\mathbf{x}_t) . 
    \end{aligned}
    \label{eq:gradient}
\end{equation}
}Therefore, the problem we need to address is transformed into modeling $p(a_{fair}=\lambda_z| \mathbf{x}_t)$.
For a sample that does not contain sensitive information, it will be uncertain to classify it definitively into any sensitive category. In other words, for debiased samples, training a classifier to predict their sensitive subgroups will result in a predicted distribution that approaches a uniform distribution. At this point, the entropy of the sensitive attribute prediction distribution $p_{\boldsymbol{\theta}_z}(z| \mathbf{x}_t)$ will reach its maximum.
Based on this property, we define 
$p(a_{fair}=\lambda_z| \mathbf{x}_t)$ as:
{
\small
\begin{equation}
\begin{aligned}
    p(a_{fair}=\lambda_z| \mathbf{x}_t)=\frac{\lambda_z H(p_{\boldsymbol{\theta}_z}(z| \mathbf{x}_t))}{C_t}, 
\end{aligned}
\label{eq:fair distribution}
\end{equation}
}where $H(\cdot)$ denotes the entropy function and $C_t$ is a normalization constant. In practice, $p_{\boldsymbol{\theta}_z}(z| \mathbf{x}_t)$ is approximated using a pre-trained sensitive classifier $\psi_{\boldsymbol{\theta}_z}$.
Adding the gradient of the logarithm of Eq.(\ref{eq:fair distribution}) in the reverse diffusion process corresponds to maximizing the entropy of $p_{\boldsymbol{\theta}_z}(z| \mathbf{x}_t)$ at each time step $t$. This ensures that the samples drawn at each step contain minimal sensitive information, making it difficult for the classifier $\psi_{\boldsymbol{\theta}_z}$ to determine their sensitive category. 
Substituting Eq.(\ref{eq:gradient}) and Eq.(\ref{eq:fair distribution}) into Eq.(\ref{eq:reverse diffusion4}) yields the final reverse SDE:
{
\small
\begin{align}
    \mathrm{d} \mathbf{x}_t=&\Big[{h}(\mathbf{x}_t)-g^2(t) (\nabla_{\mathbf{x}_t} \log p(\mathbf{x}_t)  \nonumber \\
    & + \lambda_z \nabla_{\mathbf{x}} \log H(p_{\boldsymbol{\theta}_z}(z | \mathbf{x}_t)))
    \Big] \mathrm{d}\overline{t} + g(t) \mathrm{d} \overline{\mathbf{w}},
    \label{eq:reverse diffusion5}
\end{align}
}where $\lambda_z$ is a hyperparameter that controls the guidance strength of the sensitive classifier $\psi_{\boldsymbol{\theta}_z}$. So far, we are able to generate samples that are free from sensitive information.

\textbf{Label Generation with Classifier Guidance.} To address our problem, we need to generate both the data and their labels simultaneously. Previous methods solve this by concatenating the labels with the features and then inputting them into the generative model for joint training~\cite{xu2018fairgan,van2021decaf,ramachandranpillai2023fair}. However, a fair control process can affect these approaches by influencing the generated labels, leading to undesired labels. To tackle this challenge, we propose to employ a label generation process with classifier guidance~\cite{dhariwal2021diffusion}. We first specify a label $y$ and then use the SDM to model the conditional distribution $p(\mathbf x_t| y)$. We approach this by sampling from the conditional distribution $p(\mathbf{x}_t| y)$ where $y$ represents the label condition, by solving the conditional reverse-time SDE:
{
\small
\begin{equation}
\mathrm{d} \mathbf{x}_t=\left[h(\mathbf{x}_t)-g^2(t) \nabla_{\mathbf{x}_t} \log p(\mathbf{x}| a_{fair},y)\right] \mathrm{d}\overline{t}+g(t) \mathrm{d} \overline{\mathbf{w}}.
\label{eq:reverse diffusion2}
\end{equation}
}Since $
\nabla_{\mathbf{x}_t} \log p(\mathbf{x}_t | a_{fair},y)=\nabla_{\mathbf{x}_t} \log p(a_{fair}|\mathbf{x}_t)+\nabla_{\mathbf{x}_t} \log p(\mathbf{x}_t)+\nabla_{\mathbf{x}_t} \log p(y|\mathbf{x}_t,a_{fair})$ (proved in Appendix \ref{sec:appendix-proof}), we need a pre-trained classifier to simulate $ p(y|\mathbf{x}_t,a_{fair})$. For any given sample, regardless of whether we apply fair control to it, we need to specify the generation of its label $y$ ( $y$ is independent of $a_{fair}$). Therefore, $p_{\boldsymbol{\theta}_y}(y| \mathbf{x}_t)=p_{\boldsymbol{\theta}_y}(y| \mathbf{x}_t,a_{fair})$ can be approximated using a pre-trained sensitive classifier $\phi_{\boldsymbol{\theta}_y}$.  we can rewrite Eq.(\ref{eq:reverse diffusion2}) as:
{
\small
\begin{align}
    \mathrm{d} \mathbf{x}_t=&\Big[h(\mathbf{x}_t)-g^2(t) (\nabla_{\mathbf{x}_t} \log p(\mathbf{x}_t)+ \lambda_z \nabla_{\mathbf{x}_t}\log p_{\boldsymbol{\theta}_z}(a_{fair}|\mathbf{x}_t)\nonumber\\
    &+ \lambda_y \nabla_{\mathbf{x}_t}\log p_{\boldsymbol{\theta}_y}(y|\mathbf{x}_t))\Big]\mathrm{d}\overline{t}+g(t) \mathrm{d} \overline{\mathbf{w}},
    \label{eq:reverse diffusion3}
\end{align}
}where $\lambda_y$ is a hyperparameter that controls the guidance strength of the label classifier $\phi_{\boldsymbol{\theta}_y}$.
We can specify labels for generating samples, rather than generating labels and samples simultaneously at random. Using this method, we can specify the same number of samples to be generated for different classes, thereby creating a balanced dataset to avoid insufficient feature information for minority classes in downstream classifiers. At this point, we can generate a dataset that does not contain sensitive information with labels.

\subsection{Learning Downstream Classifier
Using the Generated Data by $G$}

Let $G$ be the score network $\boldsymbol{\mathbf{\theta}}_s$ guided by $\boldsymbol{\mathbf{\theta}}_y$ and $\boldsymbol{\mathbf{\theta}}_z$. Given a target class $y$, iteratively input Gaussian noise into the generator $G$ to output a subset of data with label $y$ by Eq.(\ref{eq:reverse diffusion3}). Perform the same operation for other classes as well, and then merge all these subsets into the final dataset $\mathcal{D}^{gen}$. Our downstream training objective is to train a classifier $f_{\mathbf{\theta}}$ that satisfies:
{
\small
% \scriptsize
\begin{align}
    \min_{\boldsymbol{\theta}} \mathbb{E}_{(\mathbf{x}',z',y')\in\mathcal{D}^{gen}}[\ell(f_{\boldsymbol{\theta}}(\mathbf{x}'),y')]
\end{align}
}where $\ell$ represents a classification loss function such as cross-entropy. 
\subsection{Algorithmic Summary  of  \sysname{}}
\label{sec:appendix-algorithm}
\begin{figure}[!t]
\vspace*{-\baselineskip}
\begin{minipage}{\columnwidth}
\begin{algorithm}[H]
   \caption{\sysname{}}
   \label{alg:optimization1}
\textbf{Input}:  Labeled source datasets $\mathcal{D}^{src}$ with $S$ domains; score network $s_{\boldsymbol{\theta}_s}$; label classifier $\phi_{\boldsymbol{\theta}_y}$; sensitive classifier $\psi_{\boldsymbol{\theta}_z}$; downstream classifier $f_{\boldsymbol{\theta}}$; hyperparameters $\alpha_{s},\alpha_y,\alpha_z,\beta_{s},\beta_y,\beta_z,\gamma_{s},\gamma_y,\gamma_z$.\\
\textbf{Initialize:} $\boldsymbol{\mathbf{\theta}}_s,\boldsymbol{\mathbf{\theta}}_y,\boldsymbol{\mathbf{\theta}}_z$, and $\boldsymbol{\theta}$
\begin{algorithmic}[1]
% \State \textbf{Stage 1:}
\Repeat
\State Sample $K$ tasks $\{\mathcal{T}_k\}_{k=1}^K$ from $\mathcal{D}^{src}$
    \For{each task $\mathcal{T}_k=\{\mathcal{T}_k^{sup},\mathcal{T}_k^{qry}\}$}
        \State Evaluate $\mathcal{L}_k^{sup}$ using $\boldsymbol{\mathbf{\theta}}_s,\boldsymbol{\mathbf{\theta}}_y,\boldsymbol{\mathbf{\theta}}_z$ on $\mathcal{T}_k^{sup}$
        \State Update $\boldsymbol{\mathbf{\theta}}'_{s,k},\boldsymbol{\mathbf{\theta}}'_{y,k},\boldsymbol{\mathbf{\theta}}'_{z,k}\leftarrow\boldsymbol{\mathbf{\theta}}_s,\boldsymbol{\mathbf{\theta}}_y,\boldsymbol{\mathbf{\theta}}_z$ in Eq.(\ref{eq:where})
        \State Evaluate $\mathcal{L}_k^{qry}$ using $\boldsymbol{\mathbf{\theta}}'_{s,k},\boldsymbol{\mathbf{\theta}}'_{y,k},\boldsymbol{\mathbf{\theta}}'_{z,k}$ on $\mathcal{T}_k^{qry}$
    \EndFor
    \State Update $\boldsymbol{\mathbf{\theta}}_s,\boldsymbol{\mathbf{\theta}}_y,\boldsymbol{\mathbf{\theta}}_z$ using $\{\mathcal{L}_k^{sup},\mathcal{L}_k^{qry}\}_{k=1}^K$ in Eq.(\ref{eq:pre-training})

\Until{convergence}
% \State \textbf{Stage 2:}
\State Derive $G$ from $s_{\boldsymbol{\theta}_s}$ guided by $\phi_{\boldsymbol{\theta}_y}$ and $\psi_{\boldsymbol{\theta}_z}$ and generate $\mathcal{D}^{gen}$ from $G$ in Eq.(\ref{eq:reverse diffusion3})
% \State \textbf{Stage 3:}
\State Iteratively optimize $f_{\boldsymbol{\theta}}$ using $\mathcal{D}^{gen}$
\end{algorithmic}
\end{algorithm}
\end{minipage}

\end{figure}
The overall process of \sysname{} is illustrated in Algorithm \ref{alg:optimization1}. 
Lines 1 to 8 represent the pre-training phase of the model parameter $\Phi=\{\boldsymbol{\mathbf{\theta}}_s,\boldsymbol{\mathbf{\theta}}_y,\boldsymbol{\mathbf{\theta}}_z\}$, while lines 9 and 10 respectively represent the data generation phase and the downstream classifier training. During the pre-training phase of the model, by constructing different  $\mathcal{T}_k^{sup}$ and $\mathcal{T}_k^{qry}$, the model gains the ability to generate data from non-training domains. In the data generation phase, we do not require any additional training. We can simply use the pre-trained classifiers, $\phi_{\boldsymbol{\theta}_y} $and $\psi_{\boldsymbol{\theta}_z}$, as a guide to generate data that is free from biased information. Finally, during the downstream classifier training, only the simplest classification loss is required.
\section{Experiments}
    \label{sec:exp}
    \begin{table*}[t]
\scriptsize
\caption{Performance on \texttt{Adult} and \texttt{Bank} datasets (\textbf{bold} is the best, \underline{underline} is the second best).}
\vspace{-3mm}
\centering
\setlength\tabcolsep{2pt}
% \resizebox{0.98\textwidth}{!}{
\begin{tabular}{c|c|cccc|cccc|cccc|cccc|cccc|cccc}
\toprule

\multicolumn{26}{c}{Accuracy $\uparrow$ / $\Delta_{DP}$ $\downarrow$ / $\Delta_{EO}$ $\downarrow$ / $\Delta_{EOp}$ $\downarrow$}\\
\midrule

\parbox[t]{3mm}{\multirow{9}{*}{\rotatebox[origin=c]{90}{\texttt{Adult}}}} &
 & \multicolumn{4}{c|}{\textbf{White}} & \multicolumn{4}{c|}{\textbf{Black}} & \multicolumn{4}{c|}{\textbf{A-I-E}} & \multicolumn{4}{c|}{\textbf{A-P-I}}
 & \multicolumn{4}{c|}{\textbf{Others}} & \multicolumn{4}{c}{\textbf{Avg}}\\
\cmidrule(r){2-26}
% \specialrule{0em}{0pt}{-0.1pt}
% & Acc & $\Delta_{\text{DP}}$ & $\Delta_{\text{EO}}$ & $\Delta_{\text{EOp}}$ & Acc & $\Delta_{\text{DP}}$ & $\Delta_{\text{EO}}$ & $\Delta_{\text{EOp}}$ & Acc & $\Delta_{\text{DP}}$ & $\Delta_{\text{EO}}$ & $\Delta_{\text{EOp}}$ & Acc & $\Delta_{\text{DP}}$ & $\Delta_{\text{EO}}$ & $\Delta_{\text{EOp}}$ & Acc & $\Delta_{\text{DP}}$ & $\Delta_{\text{EO}}$ & $\Delta_{\text{EOp}}$ & Acc & $\Delta_{\text{DP}}$ & $\Delta_{\text{EO}}$ & $\Delta_{\text{EOp}}$ \\

\specialrule{0em}{0pt}{-0.1pt}

&VAE & 79.12 & 0.23 & 0.25 & 0.37 & 89.05 & 0.12 & 0.21 & 0.36 & 85.03 & 0.16 & 0.21 & 0.30 & 76.79 & 0.44 & 0.43 & 0.55 & 86.37 & 0.12 & 0.20 & 0.34 & 83.27 & 0.21 & 0.26 & 0.38 \\
&GAN & 72.27 & 0.26 & 0.27 & 0.42 & 72.07 & 0.09 & 0.33 & \textbf{0.20} & 76.51 & 0.16 & 0.22 & 0.35 & \underline{77.57} & 0.23 & 0.27 & \underline{0.22} & 83.70 & 0.11 & 0.23 & 0.25 & 76.42 & 0.17 & 0.26 & 0.29 \\
&DDPM & \underline{79.51} & 0.27 & 0.28 & 0.42 & \textbf{89.47} & 0.09 & 0.18 & 0.31 & \underline{86.74} & 0.11 & 0.17 & 0.26 & \textbf{78.37} & 0.28 & 0.33 & 0.53 & \textbf{87.34} & 0.11 & 0.20 & 0.34 & \underline{84.29} & 0.17 & 0.23 & 0.37 \\

\cmidrule(r){2-26}

&FGAN & 74.06 & \textbf{0.14} & \underline{0.17} & \textbf{0.22} & 83.44 & 0.06 & \underline{0.12} & 0.22 & 84.89 & \underline{0.06} & \textbf{0.11} & \underline{0.20} & 71.66 & \underline{0.14} & 0.19 & 0.27 & 79.91 & \textbf{0.04} & \underline{0.06} & 0.08 & 78.99 & \textbf{0.09} & \underline{0.13} & \underline{0.20} \\
&DECAF & 73.74 & 0.16 & 0.22 & 0.26 & 87.38 & 0.08 & 0.23 & 0.28 & 84.72 & \underline{0.06} & 0.26 & 0.31 & 71.99 & 0.15 & 0.24 & 0.30 & 84.26 & 0.05 & 0.12 & 0.15 & 80.42 & 0.10 & 0.21 & 0.26 \\
&FDisCo & 73.89 & 0.31 & 0.33 & 0.42 & 85.05 & \underline{0.05} & 0.19 & 0.33 & 71.49 & 0.11 & 0.23 & 0.32 & 72.43 & \textbf{0.05} & \textbf{0.12} & \textbf{0.20} & 83.15 & 0.06 & 0.07 & \textbf{0.06} & 77.20 & 0.12 & 0.19 & 0.27 \\
&FLDGM & 61.22 & 0.17 & 0.20 & 0.26 & 58.09 & 0.09 & 0.17 & 0.27 & 64.29 & 0.23 & 0.27 & 0.31 & 61.94 & 0.19 & 0.26 & 0.37 & 62.13 & 0.25 & 0.27 & 0.31 & 61.53 & 0.19 & 0.24 & 0.30 \\

\cmidrule(r){2-26}

&\sysname{} (Ours) & \textbf{81.62} & \underline{0.15} & \textbf{0.15} & \underline{0.23} & \underline{89.19} & \textbf{0.04} & \textbf{0.11} & \underline{0.21} & \textbf{87.99} & \textbf{0.05} & \underline{0.12} & \textbf{0.19} & 77.04 & 0.17 & \underline{0.18} & 0.27 & \underline{86.92} & \textbf{0.04} & \textbf{0.04} & \textbf{0.06} & \textbf{84.55} & \textbf{0.09} & \textbf{0.12} & \textbf{0.19} \\

\midrule
\midrule

\parbox[t]{2mm}{\multirow{9}{*}{\rotatebox[origin=c]{90}{\texttt{Bank}}}} &
& \multicolumn{4}{c|}{\textbf{Basic-4-years}} & \multicolumn{4}{c|}{\textbf{High-school}} & \multicolumn{4}{c|}{\textbf{Basic-9-years}} & \multicolumn{4}{c|}{\textbf{University-degree}}
 & \multicolumn{4}{c|}{\textbf{Professional-course}} & \multicolumn{4}{c}{\textbf{Avg}}\\

\cmidrule(r){2-26}
% \specialrule{0em}{0pt}{-0.1pt}
% & Acc & $\Delta_{\text{DP}}$ & $\Delta_{\text{EO}}$ & $\Delta_{\text{EOp}}$ & Acc & $\Delta_{\text{DP}}$ & $\Delta_{\text{EO}}$ & $\Delta_{\text{EOp}}$ & Acc & $\Delta_{\text{DP}}$ & $\Delta_{\text{EO}}$ & $\Delta_{\text{EOp}}$ & Acc & $\Delta_{\text{DP}}$ & $\Delta_{\text{EO}}$ & $\Delta_{\text{EOp}}$ & Acc & $\Delta_{\text{DP}}$ & $\Delta_{\text{EO}}$ & $\Delta_{\text{EOp}}$ & Acc & $\Delta_{\text{DP}}$ & $\Delta_{\text{EO}}$ & $\Delta_{\text{EOp}}$ \\

\specialrule{0em}{0pt}{-0.1pt}

&VAE & \underline{82.78} & 0.12 & 0.15 & 0.23 & \textbf{83.34} & 0.09 & 0.10 & 0.16 & \underline{85.42} & 0.07 & 0.14 & 0.24 & \textbf{79.03} & 0.08 & 0.11 & 0.17 & \textbf{83.86} & 0.05 & 0.09 & 0.14 & \textbf{82.88} & 0.08 & 0.12 & 0.19 \\
&GAN & 79.09 & 0.12 & 0.14 & 0.26 & 72.84 & 0.05 & 0.08 & 0.13 & 80.38 & 0.05 & 0.13 & 0.26 & 75.18 & 0.06 & 0.09 & 0.12 & 81.62 & 0.08 & 0.11 & 0.22 & 78.48 & 0.08 & 0.12 & 0.22 \\
&DDPM & \textbf{83.86} & 0.11 & 0.16 & 0.25 & \underline{82.71} & 0.04 & 0.10 & 0.16 & \textbf{85.58} & 0.05 & 0.13 & 0.22 & \underline{78.13} & 0.07 & 0.11 & 0.17 & \underline{82.85} & 0.04 & 0.10 & 0.19 & \underline{82.65} & 0.06 & 0.12 & 0.20 \\

\cmidrule(r){2-26}

&FGAN & 80.91 & 0.07 & 0.11 & \underline{0.18} & 78.75 & 0.03 & 0.06 & \underline{0.11} & 82.64 & 0.05 & 0.10 & 0.17 & 73.33 & \underline{0.02} & \underline{0.04} & \textbf{0.07} & 75.38 & \underline{0.04} & \textbf{0.04} & \underline{0.12} & 78.20 & \underline{0.05} & \underline{0.07} & \underline{0.13} \\
&DECAF & 80.17 & \textbf{0.06} & \underline{0.10} & 0.19 & 77.60 & \textbf{0.02} & \underline{0.05} & 0.12 & 83.24 & 0.05 & 0.09 & \underline{0.15} & 72.93 & \underline{0.02} & 0.05 & 0.13 & 71.70 & 0.12 & 0.21 & 0.13 & 77.13 & 0.06 & 0.10 & 0.14 \\
&FDisCo & 81.73 & 0.08 & 0.14 & 0.24 & 80.53 & 0.06 & 0.10 & 0.17 & 82.87 & \underline{0.04} & \underline{0.08} & 0.16 & 76.37 & 0.06 & 0.10 & 0.15 & 81.66 & 0.07 & 0.11 & 0.17 & 80.63 & 0.06 & 0.11 & 0.18 \\
&FLDGM & 75.68 & 0.17 & 0.19 & 0.21 & 74.69 & 0.07 & 0.11 & 0.15 & 78.45 & 0.08 & 0.13 & 0.19 & 70.26 & 0.08 & 0.12 & 0.15 & 73.63 & 0.07 & 0.11 & 0.15 & 74.54 & 0.10 & 0.13 & 0.17 \\

\cmidrule(r){2-26}

&\sysname{} (Ours) & 82.52 & \textbf{0.06} & \textbf{0.09} & \textbf{0.14} & 80.73 & \textbf{0.02} & \textbf{0.04} & \textbf{0.07} & 85.31 & \textbf{0.03} & \textbf{0.07} & \textbf{0.13} & 75.91 & \textbf{0.01} & \textbf{0.03} & \textbf{0.07} & 79.67 & \textbf{0.03} & \underline{0.06} & \textbf{0.11} & 80.84 & \textbf{0.03} & \textbf{0.06} & \textbf{0.11} \\
%\specialrule{0em}{1pt}{1pt}

\specialrule{0em}{0pt}{-0.1pt}
\bottomrule
\end{tabular}
% }
\label{tab:overall_result}
\vspace{-3mm}
\end{table*}

\textbf{Dataset}. We use three datasets, each containing sensitive attributes and data from different domains, to evaluate whether the data generated by the model exhibits both invariance and unbiased. 
\texttt{Adult}~\cite{kohavi1996scaling} contains a diverse set of attributes pertaining to individuals in the United States. We use the annual income as the binary label, gender as a sensitive attribute, and categorize the data into five domains based on race: White, Black, Amer-Indian-Eskimo (AIE), Asian-Pac-Islander (API), and others.
\texttt{Bank}~\cite{moro2014data} has 16 attributes and a binary label, which indicates whether the client has subscribed or not to a term deposit. We consider marital status as the binary protected attribute and partition the dataset into five domains based on five different educational levels: basic 4 years, basic 9 years, high school, university degree, and professional course. 
The \texttt{New-York-Stop-and-Frisk} (\texttt{NYSF}) dataset \cite{goel2016precinct} is a real-world dataset that includes stop, question, and frisk data for suspects across five different cities. We consider the cities as the domains from which suspects were sampled: Brooklyn, Queens, Manhattan, Bronx, and Staten Island. The suspects' gender is used as the sensitive attribute, while the target label is whether or not a suspect was frisked.

\textbf{Evaluation Metrics}. We measure the domain generalization performance using \textit{Accuracy} and evaluate the algorithmic fairness using four evaluation metrics as follows: 
\begin{itemize}[leftmargin=*]
    \item Difference of Demographic Parity ($\Delta_{DP}$)~\cite{dwork2012fairness} requires that the acceptance rate provided by the algorithm should be the same across all sensitive subgroups, which is formalized as
    \begin{equation*}
    \small
    \begin{aligned}
        \Delta_{DP}=\Big|\mathbb{P}(\hat{Y}=1|Z=0)-\mathbb{P}(\hat{Y}=1|Z=1)\Big|
    \end{aligned}
    \end{equation*}
    where $\hat{Y}$ is the predicted class label. 
    \item Difference of Equalized Odds ($\Delta_{EO}$)~\cite{hardt2016equality} examines fairness in terms of equal true positive and false positive rates, which is formalized as
    \begin{equation*}
    \small
    \begin{aligned}
        \Delta_{EO}=\frac{1}{2}\sum_y\Big|&\mathbb{P}(\hat{Y}=1|Z=0, Y=y)-\\&\mathbb{P}(\hat{Y}=1|Z=1, Y=y)\Big|
    \end{aligned}
    \end{equation*}
    \item Difference of Equalized Opportunity ($\Delta_{EOp}$)~\cite{hardt2016equality} evaluates the fairness of true positive rates specifically, which is formalized as
    % requires that $\hat{Y}$ has equal true positive rates between subgroups $Z = 0$ and $Z = 1$.
    \begin{equation*}
    \small
    \begin{aligned}
        \Delta_{EOp}=\Big|&\mathbb{P}(\hat{Y}=1|Z=0, Y=1)-\mathbb{P}(\hat{Y}=1|Z=1, Y=1)\Big|
    \end{aligned}
    \end{equation*}
    \item Difference of dataset distance ($D_{fair}$) measures the fairness of a dataset by evaluating the difference between the distances from the mean point of the generated dataset to the mean points of two sensitive groups in the original dataset. We formulally define it as:
        \begin{equation*}
    \small
    \begin{aligned}
        D_{fair} = \Big |\left \| \mu(\mathcal{D}^{gen})- \mu(\mathcal{D}_{z=0}^{src})\right \|_2-\left \|\mu(\mathcal{D}^{gen})- \mu(\mathcal{D}_{z=1}^{src}) \right \|_2 \Big |
    \end{aligned}
    \end{equation*}
    where $\mu(\cdot)$ denotes the operation of taking the mean point of a dataset, $\mathcal{D}^{src}_{z=0}$ and $\mathcal{D}^{src}_{z=1}$ represent different sensitive subsets within $\mathcal{D}^{src}$, respectively.
    % \item Ratio of Demographic Parity (DP)~\cite{dwork2012fairness} is formalized as
    % \begin{equation}
    % \mathrm{R_{DP}}= 
    % \begin{cases}
    % \displaystyle\frac{\mathbb{P}(\hat{Y}=1 \mid Z=0)}{\mathbb{P}(\hat{Y}=1 \mid Z=1)}, & \text{if } \mathrm{R_{DP}} \leq 1, \\ 
    % \displaystyle\frac{\mathbb{P}(\hat{Y}=1 \mid Z=1)}{\mathbb{P}(\hat{Y}=1 \mid Z=0)}, & \text{otherwise}.
    % \end{cases}
    % \end{equation}

    % \item Ratio of Equalized Opportunity (EOp)~\cite{hardt2016equality} is formalized as 
    % \begin{equation}
    % \mathrm{R_{EOp}}= 
    % \begin{cases}
    % \displaystyle\frac{\mathbb{P}(\hat{Y}=1 \mid Z=0, Y=1)}{\mathbb{P}(\hat{Y}=1 \mid Z=1, Y=1)} , & \text{if }\mathrm{R_{EOp}} \leq 1, \\ 
    % \displaystyle\frac{\mathbb{P}(\hat{Y}=1 \mid Z=1, Y=1)}{\mathbb{P}(\hat{Y}=1 \mid Z=0, Y=1)} , & \text{otherwise}.
    % \end{cases}
    % \end{equation}
\end{itemize}
The smaller the values of $\Delta_{DP}$,  $\Delta_{EO}$ $\Delta_{EOp}$ and $D_{fair}$, the fairer the algorithm.

\textbf{Compared Methods}. We compare \sysname{} with three classic generative models VAE~\cite{kingma2013auto}, GAN~\cite{goodfellow2020generative}, and DDPM~\cite{ho2020denoising}. Additionally, four models for fair data generation methods have also been considered: FGAN~\cite{xu2018fairgan}, DECAF~\cite{van2021decaf}, FDisCo~\cite{liu2022fair} and FLDGM~\cite{ramachandranpillai2023fair}.

\textbf{Settings}. %\textcolor{blue}{[Need to judge]}We test the performance of the model using classification tasks as an example. For the sake of fairness, we follow the same steps for training and testing all methods. Specifically, we train all generation methods under the same settings, and then train and test the classifier using the same settings.
Please refer to Appendix \ref{appendix-settings} for all hyperparameter tuning and settings, network architecture, hardware environment, training and testing strategies.
% \noindent\textbf{Model selection.} \textcolor{blue}{[Note the consistency of notations.]}We employed leave-one-domain-out cross-validation \cite{gulrajani2020search} for each methods. Specifically, given $|\mathcal{D}^s|$ training domains, we trained $|\mathcal{D}^s|$ models with the same hyperparameters, each model reserving one training domain and training on the remaining $|\mathcal{D}^s|-1$ training domains. Subsequently, each model was tested on the domain it had reserved, and the average \textit{Accuracy} across these models on their respective reserved domains was computed. The model with the highest average \textit{Accuracy} was chosen, and this model was then trained on all $|\mathcal{D}^s|$ domains.

\section{Results}
\textbf{Overall Performance.}
The overall performance of \sysname{} and its competing methods on \texttt{Adult} and \texttt{Bank} datasets is presented in Table~\ref{tab:overall_result}. Due to space limitations, we have omitted the full results with standard deviations and those for \texttt{NYSF} dataset, including them in Appendix \ref{sec:appendix-exp-overall}. Focus on the average of each metric across all domains, \sysname{} achieves the best performance in both DG and fairness on \texttt{Adult} datasets simultaneously. Although the accuracy of \sysname{} on the \texttt{Bank} dataset is lower compared to some other methods, it achieves the most equitable results with minimal performance degradation, and its accuracy is also higher than those methods that focus on fairness. Both VAE and DDPM achieve decent classification accuracy, but due to their lack of fairness consideration, they cannot guarantee the algorithmic fairness. Although FairGAN and DECAF outperform \sysname{} in fairness performance in some domains, their DG performance is not competitive. Overall, \sysname{} ensures fairness while maintaining strong classification capabilities, which means that it has successfully generated domain-invariant fair data.

% \begin{figure}[!t]
%     \centering   
%     \includegraphics[width=\columnwidth]{img/tsne_distance_old.pdf}
%     \caption{\textcolor{blue}{[Update data, x-label, y-label, legends and title in (a).]}}
%     \label{fig:exp_result_distance}
% \end{figure}
% \subsection{\textcolor{red}{Distance}}

\begin{comment}
    \begin{figure}[!t]
    \centering   
    \subfloat[LDA visualization on the White domain of \texttt{Adult} dataset.]{
        \includegraphics[width=0.45\textwidth]{img/lda_old.pdf}
        \label{fig:lda2}
    }
    \hspace{0.05\textwidth} % 添加水平间距
    \subfloat[$D_{fair}$ performance on three datasets.]{
    \setlength{\abovecaptionskip}{-10pt}
    \setlength{\belowcaptionskip}{-10pt}
        \includegraphics[width=0.45\textwidth]{img/distance.pdf}
        \label{fig:distance2}
    }
    \caption{(a) LDA visualization of the original dataset and the dataset generated by \sysname{}. The orange, green, and blue shadows represent the probability densities of the original data for two sensitive groups and the generated data, respectively. The three dashed lines indicate the means. The markers below the x-axis show the one-dimensional data, with red points representing the mean points. (b) Average $D_{fair}$ results for all fairness-aware methods across three datasets.}
    \label{fig:exp_result_distance2}
\end{figure}
\end{comment}

\begin{figure}[!t]
    \centering   
    \subfloat[LDA visualization on \texttt{Adult} dataset.]{
        \includegraphics[width=0.45\textwidth]{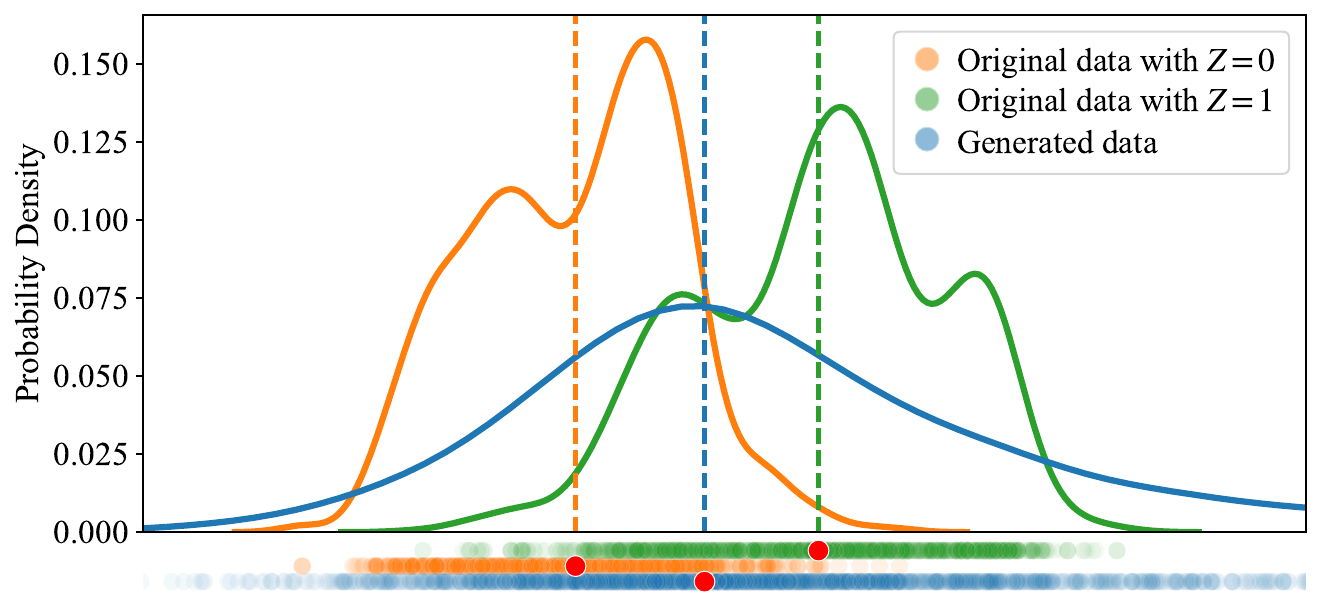}
        \vspace{-3mm}
        \label{fig:lda}
    }
    \hspace{0.05\textwidth} % 添加水平间距
    \subfloat[$D_{fair}$ performance on three datasets.]{
    \setlength{\abovecaptionskip}{-10pt}
    \setlength{\belowcaptionskip}{-10pt}
        \includegraphics[width=0.45\textwidth]{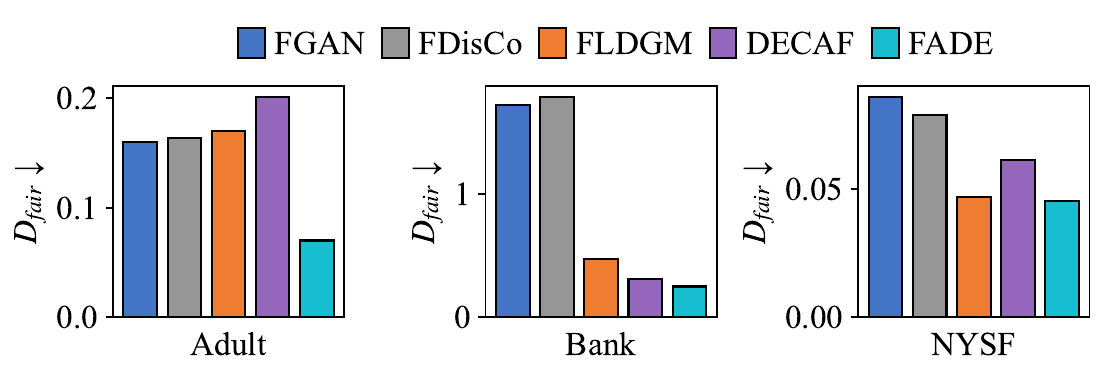}
        \vspace{-3mm}
        \label{fig:distance}
    }
    \vspace{-3mm}
    \caption{(a) LDA visualization of the original dataset and the dataset generated by \sysname{}. The orange, green, and blue curves represent the probability densities of the original data for two sensitive groups and the generated data, respectively. The three dashed lines indicate the means. The markers below the x-axis show the one-dimensional data, with red points representing the mean points. (b) Average $D_{fair}$ results for all fairness-aware methods across three datasets.}
    \label{fig:exp_result_distance}
    \vspace{-6mm}
\end{figure}

\begin{figure*}[!t]
    \centering   
    \includegraphics[width=2.1\columnwidth]{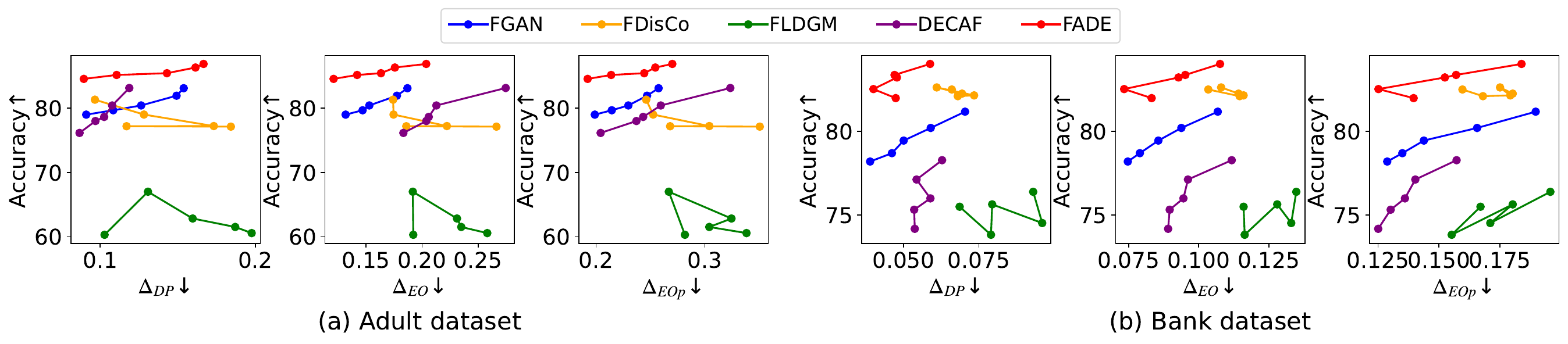}
    \vspace{-3mm}
    \caption{Accuracy-fairness trade-offs of \sysname{} by various $\lambda_z \in \{0.1, 1, 10, 50, 100\}$ on (a) \texttt{Adult} and (b) \texttt{Bank} datasets over different baselines. The upper left indicates a better trade-off performance. Results are averaged across all domains.}
    \label{fig:exp_result_tradeoff}
    \vspace{-4mm}
\end{figure*}

\textbf{Analysis of Generated Dataset.}
We further analyze the dataset generated by \sysname{} from a data perspective. We use linear discriminant analysis~\cite{balakrishnama1998linear} (LDA) to reduce the dimensionality of both the original and generated datasets to one dimension and plot their probability densities. Figure~\ref{fig:exp_result_distance}(a) shows the visualization result on the White domain of \texttt{Adult} dataset. The original data for different sensitive groups have significantly different probability density distributions and means. The mean of the data generated by \sysname{} falls between the means of the two sensitive group data, indicating that it produces a more fair data distribution. Additionally, the distribution of the generated data covers a wider range compared to the original data, suggesting that it generates generalized data beyond the original distribution.

Furthermore, we compare the average of $D_{fair}$ across all domains between the generated datasets and the original datasets for all fairness-aware methods across three datasets, as shown in Figure~\ref{fig:exp_result_distance}(b). \sysname{} achieves the best $D_{fair}$ results across three datasets, indicating that \sysname{} generates data with the smallest distance to the two original sensitive subgroups. Compared to other fairness-aware baselines, \sysname{} produced the fairest data, further validating that the fairness guidance component enables \sysname{} to generate unbiased data effectively. Meanwhile, we also observe that other methods do not perform consistently across different datasets, indicating that their debiasing performance is significantly influenced by the distribution of the original data. Particularly for DECAF, its performance is highly dependent on the causal structure of the original data. In contrast, \sysname{}'s consistent results demonstrate its strong robustness.

\textbf{Accuracy-Fairness Trade-offs.}
To evaluate the accuracy-fairness trade-offs of \sysname{}, we vary $\lambda_z \in \{0.1, 1, 10, 50, 100\}$ to obtain different classification and fairness results. A larger $\lambda_z$ indicates a greater emphasis on algorithmic fairness by \sysname{}. We compare the results with other fairness-aware methods using three fairness metrics across \texttt{Adult} and \texttt{Bank} datasets, as shown in Figure~\ref{fig:exp_result_tradeoff}. \sysname{}'s results across various $\lambda_z$ are positioned in the upper-left corner compared to all other methods. This suggests that \sysname{} not only delivers superior fairness performance but also maintains competitive classification performance, achieving the best accuracy-fairness trade-offs. Although FGAN and DECAF are comparable to \sysname{} in terms of algorithmic fairness, their lower classification accuracy results in a poorer accuracy-fairness trade-off.

\begin{table}[t]
    \centering
    \scriptsize
    \setlength\tabcolsep{2.2pt}
    \caption{Ablation study on \texttt{Adult} and \texttt{Bank} datasets. Results are averaged across all domains.}
    \vspace{-3mm}
    \begin{tabular}{llccccc}
        \toprule
          & & Accuracy$\uparrow$  & $\Delta_{DP}\downarrow$ & $\Delta_{EO}\downarrow$ & $\Delta_{EOp}\downarrow$\\
        \midrule
 \multirow{4}{*}{\texttt{Adult}} & {\sysname{}} & 84.55 & {0.089} & {0.121} & {0.192} \\
        & \textit{w/o} FG & {85.17}(\textcolor{green}{+0.62}) & 0.177 (\textcolor{red}{+0.088}) & 0.204 (\textcolor{red}{+0.083}) & 0.304 (\textcolor{red}{+0.112}) \\
        & \textit{w/o} LG & 83.18 (\textcolor{red}{-1.37}) & 0.106 (\textcolor{red}{+0.017}) & 0.138 (\textcolor{red}{+0.017}) & 0.219 (\textcolor{red}{+0.027}) \\
        & \textit{w/o} Query & 81.55 (\textcolor{red}{-3.00}) & 0.231 (\textcolor{red}{+0.142}) & 0.217 (\textcolor{red}{+0.096}) & 0.269 (\textcolor{red}{+0.077}) \\ 	 	 	 	 	 	 
       \midrule

        \multirow{4}{*}{\texttt{Bank}} & {\sysname{}} & 80.84 & {0.030} & {0.062} & {0.108} \\
        & \textit{w/o} FG & {82.45}(\textcolor{green}{+1.61}) & 0.052 (\textcolor{red}{+0.022}) & 0.104 (\textcolor{red}{+0.042}) & 0.175 (\textcolor{red}{+0.067}) \\
        & \textit{w/o} LG & 80.43 (\textcolor{red}{-0.41}) & 0.090 (\textcolor{red}{+0.060}) & 0.126 (\textcolor{red}{+0.064}) & 0.178 (\textcolor{red}{+0.070}) \\
        & \textit{w/o} Query &  80.10 (\textcolor{red}{-0.74}) & 0.032 (\textcolor{red}{+0.002}) & 0.069 (\textcolor{red}{+0.007}) & 0.111 (\textcolor{red}{+0.003}) \\ 	 	

        \bottomrule
    \end{tabular}
    \label{tab:ablation_study}
    %\vspace{-4mm}
\end{table}

\textbf{Ablation Study.}
To evaluate the effect of components in the design of \sysname{}, we conduct ablation studies as shown in Table~\ref{tab:ablation_study}. The results on the \texttt{NYPD} dataset are provided in Appendix \ref{sec:appendix-NYPD-ablation}. We propose 3 variants of \sysname{}: \textbf{(i) Without fair guidance (\textit{w/o} FG).} We set $\lambda_z$ = 0 to ensure that there is no fair control during the sampling process. Despite achieving a marginal advantage in prediction accuracy on the \texttt{adult} and \texttt{bank} dataset, a sharp increase in all three fairness metrics resulted in unfair classification outcomes (Table~\ref{tab:ablation_study}). \textbf{(ii) Without label guidance (\textit{w/o} LG).} We set $\lambda_y$ as 0 firstly. To generate the labels for the dataset, we concatenate the labels with the features and jointly train the generator using this combined input. However, by doing so, the labels generated are influenced by fair control, leading to an increase in incorrect labels and, consequently, a decline in accuracy performance. Furthermore, our empirical observations indicate that the fairness of both datasets has also decreased to varying degrees. \textbf{(iii) Without Query Set (\textit{w/o} Query).} Setting $\delta=1$ in Eq.(\ref{eq:pre-training}), the diffusion model's sampling process may retain training domain information, which can impact the accuracy in the target domain, particularly evident in the \texttt{Adult} dataset. Even if the accuracy in the \texttt{Bank} dataset does not significantly decrease, the shift in the target distribution can hinder the sensitive classifier from accurately discriminating sensitive attributes, thereby impeding the effective application of fair control and resulting in a decline in fairness.

\section{Conclusion}
    \label{sec:conclusion}
     % \textcolor{red}{[need to be revised...]} In this paper, we introduced \sysname{}: \underline{F}airness-\underline Aware \underline Diffusion with M\underline{e}ta-learning, a novel method for generating fair synthetic data from biased datasets to enhance downstream AI tasks. By using a diffusion model with gradient induction to control sample categories and obscure sensitive attributes, and training within a meta-learning framework, \sysname{} effectively addresses distribution shifts between training and test data. Experiments on real-world datasets demonstrate that \sysname{} achieves superior fairness and accuracy compared to existing methods, showcasing its potential for creating fair AI systems capable of adapting to real-world data challenges.

In this paper, we introduce \sysname{}, a three-stage framework designed to generate unbiased and domain-invariant data for fairness-aware domain generalization. By leveraging a combination of pre-trained score-based diffusion models (SDM) and classifier guidance, \sysname{} effectively eliminates bias during data generation without the need for additional training. Our approach not only enhances the generalization capabilities of the generated data but also ensures fairness when applied to downstream tasks, particularly in scenarios involving distribution shifts. Experimental results confirm that \sysname{} consistently outperforms existing methods, achieving the best accuracy-fairness trade-offs.

\clearpage
\bibliographystyle{named}
\bibliography{ijcai24}

\clearpage
\appendix
    \label{sec:appendix}
    % \section{Notation}
% \label{sec:appendix-notation}

% \section{Proof}
% \label{sec:appendix-proof}

\setcounter{figure}{5}
\setcounter{table}{2}
\onecolumn

\section{Notations}
\label{sec:appendix-notaions}

We summarize all key notations in Table~\ref{tab:appendix_notations}.

\begin{table}[h]

\caption{Important notations and their corresponding descriptions.}

\centering
\setlength{\abovecaptionskip}{0pt}
\setlength{\belowcaptionskip}{-5pt}
\small
\begin{tabular}[h]{cl}
    \toprule
    \textbf{Notations} & \textbf{Descriptions}\\
    \midrule

    $\mathcal{X}, \mathcal{Y}, \mathcal{Z}$ & Feature, output/label and sensitive space\\
    $\mathcal{E}^S, S$ & Source domains and the number of domains within it\\
    $e_i$ & A domain in $\mathcal{E}^S$ when $1\le i\le S$; a distinct shifted target domain in other cases\\
    $\mathcal{D}^{src}, \mathcal{D}^{tgt}, \mathcal{D}^{gen}, \mathcal{D}^e$ & A source dataset, a target dataset, a generated dataset and a dataset \textit{i.i.d.} sampled from a domain e\\
    $f_{\boldsymbol{\theta}}$ & A fair classifier \\
    $\omega_{\boldsymbol{\theta}}, \epsilon$ & A fair metric and a empirical threshold corresponding to it \\
    $\textbf{x}_i$ & Original data ($i=0$) and noisy data ($i>0$) in diffusion models\\
    $\beta$ & a hyperparameter to control the intensity of noise addition\\
    $G$& A generator\\
    $\mathcal{T}_k^{sup}, \mathcal{T}_k^{qry}$ & A support set and a query set in task $k$\\
    $\mathcal{B}_k^{e_i,sup}, \mathcal{B}_k^{e_i,qry}$ & Batch of domain $e_i$ in task $k$\\
    $\mathcal{L}_k^{sup}, \mathcal{L}_k^{qry}$ & Support and query losses $k$\\
    $\delta$ & An empirical weight balancing support and query losses\\
    $s_{\boldsymbol{\theta_s}}$ & Score networks\\
$\phi_{\boldsymbol{\mathbf{\theta}}_y}$ & A label classifier\\
$\psi_{\boldsymbol{\mathbf{\theta}}_z}$ & A sensitive classifier\\
$\xi_s,\xi_y,\xi_z$ & Learning rates\\
    $\mathcal{L}_s, \mathcal{L}_y, \mathcal{L}_z$ & Score loss function and cross-entropy loss functions for predicting class labels and sensitive attributes\\
    $a_{fair}$ &  A strength signal\\
    $H(\cdot)$ &  Entropy function\\
    $C_t$ &  A normalization constant\\
    $\lambda_z$ & A hyperparameter that controls the guidance strength of the sensitive classifier $\psi_{\boldsymbol{\mathbf{\theta}}_z}$\\
    $\lambda_y$ & A hyperparameter that controls the guidance strength of the label classifier $\phi_{\boldsymbol{\mathbf{\theta}}_y}$\\
    $\ell$ & A classification loss function\\

    \bottomrule 
\end{tabular}
%}
\label{tab:appendix_notations}
\end{table}

\section{Derivation of Eq.~\ref{eq:gradient} and Eq.~\ref{eq:reverse diffusion3}}
\label{sec:appendix-proof}
\subsection{Derivation of Eq.~\ref{eq:gradient}}
\begin{equation}
p(\mathbf x_t| a_{fair})= \frac{p(\mathbf x_t)p(a_{fair} | \mathbf x_t)}{p(a_{fair})}
\end{equation}
The $p(a_{fair})$ term can be treated as a constant since it does not depend on $\mathbf x_t$. Therefore, we can get $p(\mathbf x_t| a_{fair})=Z_1p(\mathbf x_t)p(a_{fair} | \mathbf x_t)$, where $Z_1$ is a constant. If we take the logarithm of both sides of the equation, we can derive as follows: 
\begin{equation}
    \begin{aligned}
     \log p(\mathbf{x}_t| a_{fair})=& \log p(\mathbf{x}_t)+ \log p(a_{fair}=|\mathbf{x}_t) +\log Z_1\\
    \nabla_{\mathbf{x}_t} \log p(\mathbf{x}_t| a_{fair})=&\nabla_{\mathbf{x}_t} \log p(\mathbf{x}_t)+\nabla_{\mathbf{x}_t} \log p(a_{fair}=|\mathbf{x}_t) . 
    \end{aligned}
\end{equation}

\subsection{Derivation of Eq.~\ref{eq:reverse diffusion3}}
\begin{equation}
p(\mathbf x_t| a_{fair},y)= \frac{p(\mathbf x_t)p(a_{fair} | \mathbf x_t)p( y| \mathbf x_t,a_{fair})}{p(a_{fair},y)}
\end{equation}
The $p(a_{fair},y)$ term can be treated as a constant since it does not depend on $\mathbf x_t$. Therefore, we can get $p(\mathbf x_t| a_{fair})=Z_2p(\mathbf x_t)p(a_{fair} | \mathbf x_t)p( y| \mathbf x_t,a_{fair})$, where $Z_2$ is a constant. If we take the logarithm of both sides of the equation, we can derive as follows: 
\begin{equation}
    \begin{aligned}
    \log p(\mathbf{x}_t | a_{fair},y)=& \log p(\mathbf{x}_t)+ \log p(a_{fair}|\mathbf{x}_t)+ \log p(y|\mathbf{x}_t,a_{fair}) +\log Z_2\\
\nabla_{\mathbf{x}_t} \log p(\mathbf{x}_t | a_{fair},y)=&\nabla_{\mathbf{x}_t} \log p(\mathbf{x}_t)+\nabla_{\mathbf{x}_t} \log p(a_{fair}|\mathbf{x}_t)+\nabla_{\mathbf{x}_t} \log p(y|\mathbf{x}_t,a_{fair})
    \end{aligned}
\end{equation}

\section{Experimental Settings}
\label{appendix-settings}
\subsection{Experimental Environments}
We programmed using the PyTorch library and ran our code on NVIDIA GeForce RTX 3090. Please refer to the compressed file for the specific code.
\subsection{Training and Testing Strategies} We employed Leave-one-domain-out cross-validation \cite{gulrajani2020search} for each methods. Specifically, given $|\mathcal{D}^s|$ training domains, we trained $|\mathcal{D}^s|$ models with the same hyperparameters, each model reserving one training domain and training on the remaining $|\mathcal{D}^{src}|-1$ training domains. Subsequently, each model was tested on the domain it had reserved, and the average performance across these models on their respective reserved domains was computed. The model with the best average performance was chosen, and this model was then trained on all $|\mathcal{D}^{src}|$ domains.
\subsection{Specific Model Architecture}
All the details of the network architecture are shown in Table.~\ref{tab:digit_encoder1} and Table.~\ref{tab:digit_encoder2}.
\label{app:architecture}
\begin{table*}[!h]
\caption{Implementation of score network $s_{\boldsymbol{\theta}_s}$. The feature dimension of the input data is $d$.}
\label{tab:digit_encoder1}
\vskip 0.15in
\begin{center}
\begin{tabular}{p{30pt}<{\raggedright} p{125pt}<{\raggedright}}

\toprule
\textbf{\#} & \textbf{Layer} \\
\midrule
1  & Linear(in=$d$, output=128) \\
2  & ReLU \\
3  & Linear(in=128, output=512) \\
4  & ReLU \\
5  & Linear(in=512, output=128) \\
6  & ReLU \\
7  & Linear(in=128, output=$d$) \\
\bottomrule
\end{tabular}
\end{center}

\end{table*}

\begin{table*}[!h]
\caption{Implementation of label classifier $\phi_{\boldsymbol{\theta}_y}$, sensitive classifier $\psi_{\boldsymbol{\theta}_z}$ and downstream classifier $f_{\boldsymbol{\theta}}$. The feature dimension of the input data is $d$.}
\label{tab:digit_encoder2}
\vskip 0.15in
\begin{center}
\begin{tabular}{p{30pt}<{\raggedright} p{125pt}<{\raggedright}}

\toprule
\textbf{\#} & \textbf{Layer} \\
\midrule
1  & Linear(in=$d$, output=128) \\
2  & ReLU \\
3  & Linear(in=128, output=512) \\
4  & ReLU \\
5  & Linear(in=512, output=128) \\
6  & ReLU \\
7  & Linear(in=128, output=2) \\
\bottomrule
\end{tabular}
\end{center}

\end{table*}

\subsection{Hyperparameter Setting and Tuning}
Table~\ref{tab:appendix_hyperparameter} shows the hyperparameter settings for each dataset. We performed a grid search to find the optimal hyperparameters within the range of [0,100] with an interval of 10.

\begin{table}[!h]
\caption{Hyperparameter settings for three datasets.}

\centering
\setlength{\abovecaptionskip}{0pt}
\setlength{\belowcaptionskip}{-5pt}
\small
\begin{tabular}[h]{c|c c}
    \toprule
    \textbf{Dataset}
    &\textbf{Hyperparameter} & \textbf{Value}\\
    \midrule
    % RL
    \multirow{2}{*}{\texttt{Adult}} & $\lambda_z$
    &100 \\
    & $\lambda_y$
    &100 \\
    & $\delta$ &0.5 \\
    \midrule
    % RL
    \multirow{2}{*}{\texttt{Bank}} & $\lambda_z$
    &30 \\
    & $\lambda_y$
    &30 \\
    & $\delta$ &0.5 \\
    \midrule

    % RL
    \multirow{2}{*}{\texttt{NYSF}} & $\lambda_z$
    &50 \\
    & $\lambda_y$
    &50 \\
    & $\delta$ &0.5 \\
    \bottomrule 
\end{tabular}
%}
\label{tab:appendix_hyperparameter}
\end{table}

\section{Detailed Experimental Results}
\label{sec:appendix-exp}
\subsection{Complete Overall Performance on \texttt{Adult}, \texttt{Bank} and \texttt{NYSF} Datasets}
\label{sec:appendix-exp-overall}
Table~\ref{tab:appendix_result_adult}, \ref{tab:appendix_result_bank} and \ref{tab:appendix_result_nysf} show DG and fairness performance on \texttt{Adult}, \texttt{Bank} and \texttt{NYSF} datasets with standard deviation.

\begin{table*}[!h]
\caption{Classification and fairness performance on \texttt{Adult} dataset with standard deviation ($\uparrow$ means higher is better, $\downarrow$ means lower is better; \textbf{bold} is the best, \underline{underline} is the second best; A-I-E stands for Amer-Indian-Eskimo, A-P-I stands for Asian-Pac-Islander).}
\centering
\scriptsize
\setlength\tabcolsep{1.5pt}
% \resizebox{0.98\textwidth}{!}{
\begin{tabular}{l|cccc|cccc|cccc}
\toprule

\multirow{2}{*}{\textbf{Method}} & \multicolumn{12}{c}{Accuracy $\uparrow$ / $\Delta_{DP}$ $\downarrow$ / $\Delta_{EO}$ $\downarrow$ / $\Delta_{EOp}$ $\downarrow$}\\
\cmidrule(r){2-13}
 & \multicolumn{4}{c|}{\textbf{White}} & \multicolumn{4}{c|}{\textbf{Black}} & \multicolumn{4}{c}{\textbf{A-I-E}}\\
\midrule

\specialrule{0em}{0pt}{-0.1pt}

VAE & 79.12\scriptsize{±0.06 }& 0.226\scriptsize{±0.002} & 0.247\scriptsize{±0.003} & 0.370\scriptsize{±0.004} & 89.05\scriptsize{±0.24} & 0.118\scriptsize{±0.001} & 0.208\scriptsize{±0.037} & 0.358\scriptsize{±0.074} & 85.03\scriptsize{±0.66} & 0.157\scriptsize{±0.019} & 0.211\scriptsize{±0.049} & 0.300\scriptsize{±0.083 }\\ 
GAN & 72.27\scriptsize{±0.01 }& 0.257\scriptsize{±0.001} & 0.266\scriptsize{±0.004} & 0.419\scriptsize{±0.003} & 72.07\scriptsize{±0.25} & 0.094\scriptsize{±0.003} & 0.327\scriptsize{±0.025} & \textbf{0.198\scriptsize{±0.009}} & 76.51\scriptsize{±0.13} & 0.162\scriptsize{±0.016} & 0.217\scriptsize{±0.054} & 0.350\scriptsize{±0.042 }\\ 
DDPM & \underline{79.51\scriptsize{±0.37}}& 0.266\scriptsize{±0.007} & 0.283\scriptsize{±0.006} & 0.417\scriptsize{±0.013} & \textbf{89.47\scriptsize{±0.07}} & 0.089\scriptsize{±0.001} & 0.177\scriptsize{±0.006} & 0.306\scriptsize{±0.014} & \underline{86.74\scriptsize{±0.07}} & 0.112\scriptsize{±0.001} & 0.172\scriptsize{±0.018} & 0.262\scriptsize{±0.045 }\\ 
\midrule
FGAN & 74.06\scriptsize{±0.13}& \textbf{0.141\scriptsize{±0.001}} & \underline{0.166\scriptsize{±0.011}} & \textbf{0.215\scriptsize{±0.004}} & 83.44\scriptsize{±0.32} & 0.061\scriptsize{±0.004} & \underline{0.124\scriptsize{±0.006}} & 0.220\scriptsize{±0.013} & 84.89\scriptsize{±0.08} & \underline{0.062\scriptsize{±0.011}} & \textbf{0.114\scriptsize{±0.007}} & \underline{0.202\scriptsize{±0.003}}\\ 
DECAF & 73.74\scriptsize{±0.06 }& 0.155\scriptsize{±0.008} & 0.218\scriptsize{±0.011} & 0.255\scriptsize{±0.002} & 87.38\scriptsize{±0.11} & 0.082\scriptsize{±0.003} & 0.228\scriptsize{±0.004} & 0.283\scriptsize{±0.008} & 84.72\scriptsize{±0.47} & 0.063\scriptsize{±0.003} & 0.258\scriptsize{±0.006} & 0.309\scriptsize{±0.016 }\\ 
FDisCo & 73.89\scriptsize{±0.02 }& 0.314\scriptsize{±0.016} & 0.327\scriptsize{±0.014} & 0.422\scriptsize{±0.005} & 85.05\scriptsize{±0.15} & \underline{0.054\scriptsize{±0.001}} & 0.189\scriptsize{±0.014} & 0.332\scriptsize{±0.029} & 71.49\scriptsize{±0.91} & 0.111\scriptsize{±0.004} & 0.232\scriptsize{±0.030} & 0.323\scriptsize{±0.051 }\\ 
FLDGM & 61.22\scriptsize{±5.11 }& 0.173\scriptsize{±0.073} & 0.202\scriptsize{±0.052} & 0.262\scriptsize{±0.037} & 58.09\scriptsize{±0.11} & 0.091\scriptsize{±0.005} & 0.174\scriptsize{±0.028} & 0.267\scriptsize{±0.046} & 64.29\scriptsize{±0.02} & 0.235\scriptsize{±0.005} & 0.268\scriptsize{±0.006} & 0.314\scriptsize{±0.015 }\\ 
\midrule
\sysname{} & \textbf{81.62\scriptsize{±0.16 }}& \underline{0.151\scriptsize{±0.003}} & \textbf{0.149\scriptsize{±0.001}} & \underline{0.230\scriptsize{±0.003}} & \underline{89.19\scriptsize{±0.16}} & \textbf{0.039\scriptsize{±0.004}} & \textbf{0.112\scriptsize{±0.000}} & \underline{0.208\scriptsize{±0.028}} & \textbf{87.99\scriptsize{±0.32}} & \textbf{0.048\scriptsize{±0.003}} & \underline{0.116\scriptsize{±0.011}} & \textbf{0.187\scriptsize{±0.011 }}\\  
%\specialrule{0em}{1pt}{1pt}

\cmidrule(r){1-13}
 & \multicolumn{4}{c|}{\textbf{A-P-I}} & \multicolumn{4}{c|}{\textbf{Others}} & \multicolumn{4}{c}{\textbf{Avg}}\\
\midrule

\specialrule{0em}{0pt}{-0.1pt}

VAE & 76.79\scriptsize{±0.20} & 0.437\scriptsize{±0.033} & 0.431\scriptsize{±0.069} & 0.546\scriptsize{±0.095} & 86.37\scriptsize{±1.88} & 0.125\scriptsize{±0.046} & 0.203\scriptsize{±0.014} & 0.337\scriptsize{±0.021} & 83.27 & 0.211 & 0.260 & 0.382 \\ 
GAN & \underline{77.57\scriptsize{±1.09}} & 0.235\scriptsize{±0.032} & 0.266\scriptsize{±0.030} & \underline{0.215\scriptsize{±0.027}} & 83.70\scriptsize{±1.24} & 0.107\scriptsize{±0.010} & 0.228\scriptsize{±0.007} & 0.252\scriptsize{±0.004} & 76.42 & 0.171 & 0.261 & 0.287 \\ 
DDPM & \textbf{78.37\scriptsize{±2.43}} & 0.281\scriptsize{±0.007} & 0.334\scriptsize{±0.025} & 0.525\scriptsize{±0.062} & \textbf{87.34\scriptsize{±0.23}} & 0.109\scriptsize{±0.011} & 0.202\scriptsize{±0.010} & 0.337\scriptsize{±0.017} & \underline{84.29} & 0.171 & 0.234 & 0.369 \\  
\midrule
FGAN & 71.66\scriptsize{±0.13} & \underline{0.142\scriptsize{±0.001}} & 0.191\scriptsize{±0.001} & 0.269\scriptsize{±0.001} & 79.91\scriptsize{±1.15} & \underline{0.041\scriptsize{±0.004}} & \underline{0.064\scriptsize{±0.004}} & 0.078\scriptsize{±0.002} & 78.99 & \underline{0.091} & \underline{0.132} & \underline{0.199} \\ 
DECAF & 71.99\scriptsize{±0.35} & 0.145\scriptsize{±0.006} & 0.242\scriptsize{±0.006} & 0.298\scriptsize{±0.003} & 84.26\scriptsize{±0.69} & 0.045\scriptsize{±0.005} & 0.118\scriptsize{±0.024} & 0.152\scriptsize{±0.015} & 80.42 & 0.098 & 0.213 & 0.259 \\ 
FDisCo & 72.43\scriptsize{±0.35} & \textbf{0.046\scriptsize{±0.016}} & \textbf{0.118\scriptsize{±0.004}} & \textbf{0.204\scriptsize{±0.017}} & 83.15\scriptsize{±2.75} & 0.061\scriptsize{±0.038} & 0.069\scriptsize{±0.019} & \underline{0.064\scriptsize{±0.003}} & 77.20 & 0.117 & 0.187 & 0.269 \\ 
FLDGM & 61.94\scriptsize{±0.86} & 0.189\scriptsize{±0.022} & 0.262\scriptsize{±0.011} & 0.374\scriptsize{±0.026} & 62.13\scriptsize{±3.13} & 0.252\scriptsize{±0.023} & 0.272\scriptsize{±0.085} & 0.306\scriptsize{±0.168} & 61.53 & 0.188 & 0.235 & 0.304 \\ 
\midrule
\sysname{} & 77.04\scriptsize{±2.52} & 0.170\scriptsize{±0.028} & \underline{0.183\scriptsize{±0.022}} & 0.273\scriptsize{±0.037} & \underline{86.92\scriptsize{±0.69}} & \textbf{0.040\scriptsize{±0.024}} & \textbf{0.044\scriptsize{±0.021}} & \textbf{0.063\scriptsize{±0.014}} & \textbf{84.55} & \textbf{0.089} & \textbf{0.121} & \textbf{0.192} \\ 
%\specialrule{0em}{1pt}{1pt}

\specialrule{0em}{0pt}{-0.1pt}
\bottomrule
\end{tabular}
% }
\label{tab:appendix_result_adult}
\end{table*}

\begin{table*}[h]
\caption{Classification and fairness performance on \texttt{Bank} dataset with standard deviation ($\uparrow$ means higher is better, $\downarrow$ means lower is better; \textbf{bold} is the best, \underline{underline} is the second best).}
\centering
\scriptsize
\setlength\tabcolsep{1.5pt}
% \resizebox{0.98\textwidth}{!}{
\begin{tabular}{l|cccc|cccc|cccc}
\toprule
 
\multirow{2}{*}{\textbf{Method}} & \multicolumn{12}{c}{Accuracy $\uparrow$ / $\Delta_{DP}$ $\downarrow$ / $\Delta_{EO}$ $\downarrow$ / $\Delta_{EOp}$ $\downarrow$}\\
\cmidrule(r){2-13}
 & \multicolumn{4}{c|}{\textbf{Basic-4-years}} & \multicolumn{4}{c|}{\textbf{High-school}} & \multicolumn{4}{c}{\textbf{Basic-9-years}}\\
\midrule

\specialrule{0em}{0pt}{-0.1pt}

VAE& \underline{82.78\scriptsize{±0.06}} & 0.121\scriptsize{±0.001} & 0.151\scriptsize{±0.002} & 0.226\scriptsize{±0.006 }& \textbf{83.34\scriptsize{±0.02 }}& 0.086\scriptsize{±0.037} & 0.104\scriptsize{±0.011} & 0.157\scriptsize{±0.030} & \underline{85.42\scriptsize{±0.02}} & 0.067\scriptsize{±0.008} & 0.143\scriptsize{±0.003} & 0.245\scriptsize{±0.006 }\\ 
GAN& 79.09\scriptsize{±0.75} & 0.118\scriptsize{±0.007} & 0.143\scriptsize{±0.001} & 0.255\scriptsize{±0.003 }& 72.84\scriptsize{±0.54 }& 0.050\scriptsize{±0.006} & 0.080\scriptsize{±0.005} & 0.128\scriptsize{±0.006} & 80.38\scriptsize{±1.17} & 0.055\scriptsize{±0.001} & 0.133\scriptsize{±0.001} & 0.262\scriptsize{±0.008 }\\ 
DDPM& \textbf{83.86\scriptsize{±0.02}} & 0.111\scriptsize{±0.007} & 0.155\scriptsize{±0.005} & 0.250\scriptsize{±0.013 }& \underline{82.71\scriptsize{±0.20 }}& 0.044\scriptsize{±0.007} & 0.098\scriptsize{±0.002} & 0.160\scriptsize{±0.001} & \textbf{85.58\scriptsize{±0.12}} & 0.047\scriptsize{±0.006} & 0.129\scriptsize{±0.004} & 0.217\scriptsize{±0.003 }\\ 
\midrule
FGAN& 80.91\scriptsize{±0.45} & 0.067\scriptsize{±0.001} & 0.106\scriptsize{±0.005} & \underline{0.181\scriptsize{±0.008 }}& 78.75\scriptsize{±0.11 }& 0.033\scriptsize{±0.003} & 0.060\scriptsize{±0.005} & \underline{0.109\scriptsize{±0.003}} & 82.64\scriptsize{±0.69} & 0.049\scriptsize{±0.005} & 0.098\scriptsize{±0.014} & 0.166\scriptsize{±0.017 }\\ 
DECAF& 80.17\scriptsize{±0.07} & \underline{0.062\scriptsize{±0.001}} & \underline{0.099\scriptsize{±0.004}} & 0.188\scriptsize{±0.004 }& 77.60\scriptsize{±0.06 }& \underline{0.022\scriptsize{±0.001}} & \underline{0.047\scriptsize{±0.001}} & 0.123\scriptsize{±0.002} & 83.24\scriptsize{±0.12} & 0.049\scriptsize{±0.003} & 0.087\scriptsize{±0.001} & \underline{0.150\scriptsize{±0.003 }}\\ 
FDisCo& 81.73\scriptsize{±0.27} & 0.077\scriptsize{±0.011} & 0.144\scriptsize{±0.022} & 0.240\scriptsize{±0.041 }& 80.53\scriptsize{±0.49 }& 0.055\scriptsize{±0.008} & 0.102\scriptsize{±0.013} & 0.169\scriptsize{±0.021} & 82.87\scriptsize{±0.57} & \underline{0.040\scriptsize{±0.006}} & \underline{0.081\scriptsize{±0.005}} & 0.147\scriptsize{±0.017 }\\ 
FLDGM& 75.68\scriptsize{±0.25} & 0.172\scriptsize{±0.029} & 0.191\scriptsize{±0.032} & 0.213\scriptsize{±0.037 }& 74.69\scriptsize{±0.58 }& 0.073\scriptsize{±0.002} & 0.113\scriptsize{±0.004} & 0.152\scriptsize{±0.005} & 78.45\scriptsize{±0.80} & 0.084\scriptsize{±0.011} & 0.129\scriptsize{±0.003} & 0.191\scriptsize{±0.016 }\\ 
\midrule
\sysname{}& 82.52\scriptsize{±0.27} & \textbf{0.058\scriptsize{±0.003}} & \textbf{0.088\scriptsize{±0.004}} & \textbf{0.139\scriptsize{±0.011 }}& 80.73\scriptsize{±0.57 }& \textbf{0.020\scriptsize{±0.003}} & \textbf{0.042\scriptsize{±0.009}} & \textbf{0.073\scriptsize{±0.018}} & 85.31\scriptsize{±0.21} & \textbf{0.029\scriptsize{±0.005}} & \textbf{0.067\scriptsize{±0.005}} & \textbf{0.129\scriptsize{±0.008 }}\\   
%\specialrule{0em}{1pt}{1pt}

\cmidrule(r){1-13}
 & \multicolumn{4}{c|}{\textbf{University-degree}} & \multicolumn{4}{c|}{\textbf{Professional-course}} & \multicolumn{4}{c}{\textbf{Avg}}\\
\midrule

\specialrule{0em}{0pt}{-0.1pt}

VAE & \textbf{79.03\scriptsize{±0.11}} & 0.080\scriptsize{±0.002} & 0.112\scriptsize{±0.011} & 0.167\scriptsize{±0.016} & \textbf{83.86\scriptsize{±0.18}} & 0.054\scriptsize{±0.004} & 0.089\scriptsize{±0.001} & 0.137\scriptsize{±0.001} & \textbf{82.88} & 0.082 & 0.120 & 0.186 \\ 
GAN & 75.18\scriptsize{±0.47} & 0.059\scriptsize{±0.006} & 0.087\scriptsize{±0.002} & 0.124\scriptsize{±0.001} & 81.62\scriptsize{±0.56} & 0.082\scriptsize{±0.006} & 0.114\scriptsize{±0.002} & 0.217\scriptsize{±0.006} & 78.48 & 0.076 & 0.117 & 0.215 \\ 
DDPM & \underline{78.13\scriptsize{±0.16}} & 0.065\scriptsize{±0.004} & 0.110\scriptsize{±0.007} & 0.168\scriptsize{±0.011} & \underline{82.85\scriptsize{±0.26}} & 0.043\scriptsize{±0.007} & 0.104\scriptsize{±0.016} & 0.187\scriptsize{±0.012} & \underline{82.63} & 0.062 & 0.119 & 0.196 \\ 
\midrule
FGAN & 73.33\scriptsize{±0.17} & 0.019\scriptsize{±0.003} & \underline{0.037\scriptsize{±0.003}} & \underline{0.071\scriptsize{±0.008}} & 75.38\scriptsize{±1.63} & \underline{0.043\scriptsize{±0.004}} & \textbf{0.038\scriptsize{±0.008}} & \underline{0.122\scriptsize{±0.001}} & 78.20 & \underline{0.046} & \underline{0.065} & \underline{0.129} \\ 
DECAF & 72.93\scriptsize{±0.59} & \underline{0.019\scriptsize{±0.001}} & 0.049\scriptsize{±0.002} & 0.131\scriptsize{±0.005} & 71.70\scriptsize{±0.64} & 0.121\scriptsize{±0.142} & 0.205\scriptsize{±0.239} & 0.131\scriptsize{±0.002} & 77.13 & 0.058 & 0.096 & 0.140 \\ 
FDisCo & 76.37\scriptsize{±0.69} & 0.059\scriptsize{±0.008} & 0.099\scriptsize{±0.003} & 0.146\scriptsize{±0.008} & 81.66\scriptsize{±0.50} & 0.068\scriptsize{±0.005} & 0.107\scriptsize{±0.004} & 0.175\scriptsize{±0.006} & 80.63 & 0.061 & 0.108 & 0.175 \\ 
FLDGM & 70.26\scriptsize{±0.75} & 0.081\scriptsize{±0.004} & 0.118\scriptsize{±0.003} & 0.153\scriptsize{±0.008} & 73.63\scriptsize{±1.24} & 0.071\scriptsize{±0.023} & 0.113\scriptsize{±0.004} & 0.148\scriptsize{±0.013} & 74.54 & 0.096 & 0.133 & 0.171 \\ 
\midrule
\sysname{} & 75.91\scriptsize{±0.68} & \textbf{0.014\scriptsize{±0.001}} & \textbf{0.033\scriptsize{±0.005}} & \textbf{0.068\scriptsize{±0.009}} & 79.67\scriptsize{±1.44} & \textbf{0.026\scriptsize{±0.003}} & \underline{0.062\scriptsize{±0.015}} & \textbf{0.111\scriptsize{±0.026}} & 80.84 & \textbf{0.030} & \textbf{0.062} & \textbf{0.108} \\ 
%\specialrule{0em}{1pt}{1pt}

\specialrule{0em}{0pt}{-0.1pt}
\bottomrule
\end{tabular}
% }
\label{tab:appendix_result_bank}
\end{table*}

\begin{table*}[h]
\caption{Classification and fairness performance on \texttt{NYSF} dataset with standard deviation ($\uparrow$ means higher is better, $\downarrow$ means lower is better; \textbf{bold} is the best, \underline{underline} is the second best).}
\centering
\scriptsize
\setlength\tabcolsep{1.5pt}
% \resizebox{0.98\textwidth}{!}{
\begin{tabular}{l|cccc|cccc|cccc}
\toprule
 
\multirow{2}{*}{\textbf{Method}} & \multicolumn{12}{c}{Accuracy $\uparrow$ / $\Delta_{DP}$ $\downarrow$ / $\Delta_{EO}$ $\downarrow$ / $\Delta_{EOp}$ $\downarrow$}\\
\cmidrule(r){2-13}
 & \multicolumn{4}{c|}{\textbf{Brooklyn}} & \multicolumn{4}{c|}{\textbf{Queens}} & \multicolumn{4}{c}{\textbf{Manhattan}}\\
\midrule

\specialrule{0em}{0pt}{-0.1pt}

VAE& \underline{54.85\scriptsize{±0.06}} & 0.147\scriptsize{±0.006} & 0.228\scriptsize{±0.019} & 0.299\scriptsize{±0.004 }& 58.29\scriptsize{±3.02 }& 0.180\scriptsize{±0.020} & 0.286\scriptsize{±0.001} & 0.327\scriptsize{±0.004} & 58.04\scriptsize{±0.04} & 0.166\scriptsize{±0.003} & 0.233\scriptsize{±0.005} & 0.297\scriptsize{±0.013 }\\ 
GAN& 51.15\scriptsize{±0.04} & 0.165\scriptsize{±0.004} & 0.186\scriptsize{±0.004} & 0.302\scriptsize{±0.001 }& 54.44\scriptsize{±0.33 }& 0.198\scriptsize{±0.004} & 0.284\scriptsize{±0.001} & 0.300\scriptsize{±0.002} & 56.18\scriptsize{±0.08} & 0.151\scriptsize{±0.001} & \underline{0.225\scriptsize{±0.001}} & \textbf{0.281\scriptsize{±0.004 }} \\ 
DDPM& 54.70\scriptsize{±0.45} & 0.140\scriptsize{±0.002} & 0.180\scriptsize{±0.004} & 0.245\scriptsize{±0.001 }& \textbf{61.51\scriptsize{±0.15 }}& 0.194\scriptsize{±0.003} & 0.273\scriptsize{±0.000} & 0.335\scriptsize{±0.001} & \textbf{60.20\scriptsize{±0.53}} & 0.169\scriptsize{±0.007} & 0.228\scriptsize{±0.005} & 0.301\scriptsize{±0.009 }\\  
\midrule
FGAN& 52.71\scriptsize{±0.71} & \underline{0.110\scriptsize{±0.003}} & \textbf{0.171\scriptsize{±0.002}} & 0.223\scriptsize{±0.001 }& 54.28\scriptsize{±0.09 }& \textbf{0.155\scriptsize{±0.004}} & 0.249\scriptsize{±0.003} & 0.308\scriptsize{±0.006} & 57.72\scriptsize{±0.57} & 0.148\scriptsize{±0.008} & 0.240\scriptsize{±0.003} & 0.307\scriptsize{±0.008 }\\ 
DECAF& 52.64\scriptsize{±0.69} & 0.121\scriptsize{±0.001} & 0.220\scriptsize{±0.002} & \underline{0.242\scriptsize{±0.002 }}& 56.24\scriptsize{±0.04 }& 0.170\scriptsize{±0.003} & 0.241\scriptsize{±0.001} & 0.313\scriptsize{±0.013} & 57.00\scriptsize{±0.18} & \underline{0.147\scriptsize{±0.006}} & 0.233\scriptsize{±0.003} & \underline{0.285\scriptsize{±0.004 }}\\ 
FDisCo& 53.31\scriptsize{±0.17} & 0.170\scriptsize{±0.001} & 0.261\scriptsize{±0.001} & 0.312\scriptsize{±0.001 }& 45.98\scriptsize{±0.45 }& 0.168\scriptsize{±0.030} & \underline{0.240\scriptsize{±0.010}} & \textbf{0.265\scriptsize{±0.025}} & 52.60\scriptsize{±0.20} & 0.158\scriptsize{±0.000} & 0.243\scriptsize{±0.001} & 0.288\scriptsize{±0.001 }\\ 
FLDGM& 53.70\scriptsize{±0.15} & 0.150\scriptsize{±0.003} & 0.246\scriptsize{±0.003} & 0.305\scriptsize{±0.001 }& \underline{58.41\scriptsize{±1.26 }}& 0.166\scriptsize{±0.005} & 0.273\scriptsize{±0.003} & 0.315\scriptsize{±0.001} & 53.56\scriptsize{±0.19} & \textbf{0.144\scriptsize{±0.013}} & 0.229\scriptsize{±0.002} & 0.287\scriptsize{±0.001 }\\ 
\midrule
\sysname{}& \textbf{56.90\scriptsize{±0.33}} & \textbf{0.109\scriptsize{±0.006}} & \underline{0.173\scriptsize{±0.004}} & \textbf{0.224\scriptsize{±0.008}}& 55.78\scriptsize{±0.06}& \underline{0.166\scriptsize{±0.001}} & \textbf{0.240\scriptsize{±0.005}} & \underline{0.297\scriptsize{±0.006}} & \underline{60.11\scriptsize{±0.11}} & 0.152\scriptsize{±0.004} & \textbf{0.224\scriptsize{±0.002}} & 0.296\scriptsize{±0.003 }\\    
%\specialrule{0em}{1pt}{1pt}

\cmidrule(r){1-13}
 & \multicolumn{4}{c|}{\textbf{Bronx}} & \multicolumn{4}{c|}{\textbf{Staten island}} & \multicolumn{4}{c}{\textbf{Avg}}\\
\midrule

\specialrule{0em}{0pt}{-0.1pt}

VAE & \underline{56.26\scriptsize{±0.13}} & 0.161\scriptsize{±0.011} & 0.232\scriptsize{±0.001} & 0.248\scriptsize{±0.001} & 57.59\scriptsize{±0.04} & 0.148\scriptsize{±0.009} & 0.202\scriptsize{±0.006} & 0.258\scriptsize{±0.001} & 57.00 & 0.160 & 0.236 & 0.286 \\ 
GAN & 53.68\scriptsize{±0.64} & 0.160\scriptsize{±0.011} & 0.223\scriptsize{±0.001} & \textbf{0.199\scriptsize{±0.075}} & 57.89\scriptsize{±0.10} & 0.136\scriptsize{±0.002} & 0.179\scriptsize{±0.001} & 0.251\scriptsize{±0.011} & 54.67 & 0.162 & 0.218 & 0.266 \\ 
DDPM & \textbf{56.40\scriptsize{±0.21}} & 0.170\scriptsize{±0.001} & 0.231\scriptsize{±0.005} & 0.252\scriptsize{±0.010} & \underline{62.60\scriptsize{±0.05}} & 0.158\scriptsize{±0.001} & 0.196\scriptsize{±0.003} & 0.258\scriptsize{±0.004} & \textbf{59.08} & 0.166 & 0.221 & 0.278 \\ 
\midrule
FGAN & 53.68\scriptsize{±0.66} & 0.163\scriptsize{±0.001} & 0.236\scriptsize{±0.008} & 0.244\scriptsize{±0.003} & 58.58\scriptsize{±0.43} & 0.139\scriptsize{±0.001} & 0.179\scriptsize{±0.004} & \textbf{0.211\scriptsize{±0.001}} & 55.39 & \underline{0.143} & \underline{0.215} & \underline{0.258} \\ 
DECAF & 52.35\scriptsize{±0.28} & 0.152\scriptsize{±0.004} & 0.218\scriptsize{±0.006} & 0.240\scriptsize{±0.001} & 59.52\scriptsize{±0.43} & 0.133\scriptsize{±0.001} & 0.177\scriptsize{±0.003} & 0.227\scriptsize{±0.008} & 55.55 & 0.145 & 0.218 & 0.261 \\ 
FDisCo & 46.11\scriptsize{±0.92} & \textbf{0.139\scriptsize{±0.011}} & \underline{0.211\scriptsize{±0.010}} & \underline{0.218\scriptsize{±0.008}} & 54.78\scriptsize{±0.16} & \textbf{0.101\scriptsize{±0.006}} & \underline{0.170\scriptsize{±0.004}} & \underline{0.218\scriptsize{±0.002}} & 50.55 & 0.147 & 0.225 & 0.260 \\ 
FLDGM & 55.42\scriptsize{±3.53} & 0.144\scriptsize{±0.009} & 0.223\scriptsize{±0.009} & 0.234\scriptsize{±0.005} & 53.84\scriptsize{±0.19} & \underline{0.110\scriptsize{±0.006}} & 0.173\scriptsize{±0.000} & 0.221\scriptsize{±0.006} & 54.98 & \underline{0.143} & 0.229 & 0.272 \\ 
\midrule
\sysname{} & 51.26\scriptsize{±0.86} & \underline{0.144\scriptsize{±0.002}} & \textbf{0.201\scriptsize{±0.008}} & 0.229\scriptsize{±0.007} & \textbf{66.31\scriptsize{±0.08}} & 0.130\scriptsize{±0.009} & \textbf{0.170\scriptsize{±0.002}} & 0.227\scriptsize{±0.009} & \underline{58.07} & \textbf{0.140} & \textbf{0.202} & \textbf{0.255} \\ 
%\specialrule{0em}{1pt}{1pt}

\specialrule{0em}{0pt}{-0.1pt}
\bottomrule
\end{tabular}
% }
\label{tab:appendix_result_nysf}
\end{table*}

\subsection{Ablation Study on \texttt{NYSF} Dataset}
\label{sec:appendix-NYPD-ablation}
Table~\ref{tab:ablation_study_nysf} shows ablation study performance on \texttt{NYSF} dataset with standard deviation.

\begin{table}[h]
    \centering
    \scriptsize
    \caption{Ablation study on \texttt{NYSF}. Results are averaged across all domains. ($\uparrow$ means higher is better, $\downarrow$ means lower is better; \textbf{bold} is the best.)}
    \begin{tabular}{lccccc}
        \toprule
  \multirow{4}{*}{\texttt{NYSF}} &{\sysname{}} & 58.07 & {0.140} & {0.202 } & 0.255 \\
        &\textit{w/o} FG & {60.69}(\textcolor{green}{+2.62}) & 0.170 (\textcolor{red}{+0.030}) & 0.229 (\textcolor{red}{+0.027}) & 0.287 (\textcolor{red}{+0.032}) \\
        &\textit{w/o} LG & 59.96 (\textcolor{green}{+1.89}) & 0.162 (\textcolor{red}{+0.022}) & 0.222 (\textcolor{red}{+0.020}) & {0.248} (\textcolor{green}{-0.007}) \\
        &\textit{w/o} Query &  57.60 (\textcolor{red}{-0.47}) & 0.142 (\textcolor{red}{+0.002}) & 0.203 (\textcolor{red}{+0.001}) & 0.259 (\textcolor{red}{+0.004}) \\ 

        \bottomrule
    \end{tabular}
    \label{tab:ablation_study_nysf}
\end{table}

\end{document}